\documentclass[pdflatex, sn-basic, iicol]{sn-jnl}

\usepackage{graphicx}%
\usepackage{multirow}%
\usepackage{amsmath,amssymb,amsfonts}%
\usepackage{amsthm}%
\usepackage{mathrsfs}%
\usepackage[title]{appendix}%
\usepackage{textcomp}%
\usepackage{manyfoot}%
\usepackage{booktabs}%
\usepackage{algorithm}%
\usepackage{algorithmicx}%
\usepackage{algpseudocode}%
\usepackage{listings}%
\usepackage{makecell}
\usepackage{lmodern}
\usepackage{mathrsfs}

\usepackage[dvipsnames]{xcolor}

\usepackage{xspace}
\usepackage{gensymb}
\usepackage{multirow}
\usepackage{arydshln}
\usepackage[normalem]{ulem}
\usepackage[none]{hyphenat}
\usepackage{caption}

\def\eg{\emph{e.g.}\xspace}
\def\ie{\emph{i.e.}\xspace}

\newcommand{\EventEgo}{\mbox{EventEgo3D}}
\newcommand{\EventEgoPP}{\mbox{EventEgo3D++}}

\begin{document}

\title[Article Title]{EventEgo3D++: 3D Human Motion Capture from 
a~Head-Mounted Event Camera}

\author[1,2]{\fnm{Christen} \sur{Millerdurai}}\email{christen.millerdurai@dfki.de}
\author[1]{\fnm{Hiroyasu} \sur{Akada}}\email{hakada@mpi-inf.mpg.de}
\author[1]{\fnm{Jian} \sur{Wang}}\email{jianwang@mpi-inf.mpg.de}
\author[1]{\fnm{Diogo} \sur{Luvizon}}\email{dluvizon@mpi-inf.mpg.de}
\author[2]{\fnm{Alain} \sur{Pagani}}\email{alain.pagani@dfki.de}
\author[2]{\fnm{Didier} \sur{Stricker}}\email{didier.stricker@dfki.de}
\author[1]{\fnm{Christian} \sur{Theobalt}}\email{theobalt@mpi-inf.mpg.de}
\author[1]{\fnm{Vladislav} \sur{Golyanik}}\email{golyanik@mpi-inf.mpg.de}

\affil[1]{\orgdiv{Visual Computing and Artificial Intelligence}, \orgname{Max Planck Institute for Informatics, SIC}, \orgaddress{\street{Stuhlsatzenhausweg E1 4}, \city{Saarbrücken}, \postcode{66123}, \state{Saarland}, \country{Germany}}}

\affil[2]{\orgdiv{Augmented Vision}, \orgname{German Research Center for Artificial Intelligence (DFKI)}, \orgaddress{\street{Trippstadter Str. 122}, \city{Kaiserslautern}, \postcode{67663}, \state{Rhineland-Palatinate}, \country{Germany}}}

\abstract{
Monocular egocentric 3D human motion capture remains a significant challenge, particularly under conditions of low lighting and fast movements, which are common in head-mounted device applications.
Existing methods that rely on RGB cameras often fail under these conditions. 
To address these limitations, we introduce EventEgo3D++, the first approach that leverages a monocular event camera with a fisheye lens for 3D human motion capture. 
Event cameras excel in high-speed scenarios and varying illumination due to their high temporal resolution, providing reliable cues for accurate 3D human motion capture.
EventEgo3D++ leverages the LNES representation of event streams to enable precise 3D reconstructions.
We have also developed a mobile head-mounted device (HMD) prototype equipped with an event camera, capturing a comprehensive dataset that includes real event observations from both controlled studio environments and in-the-wild settings, in addition to a synthetic dataset.
Additionally, to provide a more holistic dataset, we include allocentric RGB streams that offer different perspectives of the HMD wearer, along with their corresponding SMPL body model.
Our experiments demonstrate that EventEgo3D++ achieves superior 3D accuracy and robustness compared to existing solutions, even in challenging conditions. 
Moreover, our method supports real-time 3D pose updates at a rate of 140Hz. 
This work is an extension of the EventEgo3D approach (CVPR 2024) and further advances the state of the art in egocentric 3D human motion capture. For more details, visit the project page at \url{https://eventego3d.mpi-inf.mpg.de}.
}


\maketitle

\section{Introduction}\label{sec:intro} 
Head-mounted devices (HMDs) hold significant potential to become the next major platform for mobile and pervasive computing, offering diverse applications in many fields such as education, driving, personal assistance systems, and gaming. 
HMDs enhance user flexibility, allowing individuals to move freely and explore their surroundings seamlessly. 
As a result, egocentric 3D human pose estimation has emerged as an active research area, with numerous studies focusing on recovering 3D human poses using down-facing fisheye RGB cameras mounted on HMDs~\citep{rhodin2016egocap, xu2019mo2cap2, zhao2021egoglass, wang2022estimating, hakada2022unrealego, hakada2024unrealego2, wang2023scene, wang2021estimating, Tom2023SelfPose3E, Liu2023, Li2023EgoBody, wang2024egocentric, Kang2023Ego3DPose, kang2024egotap}.

\begin{figure*}[t]
\begin{center}
    \includegraphics[width=\linewidth]{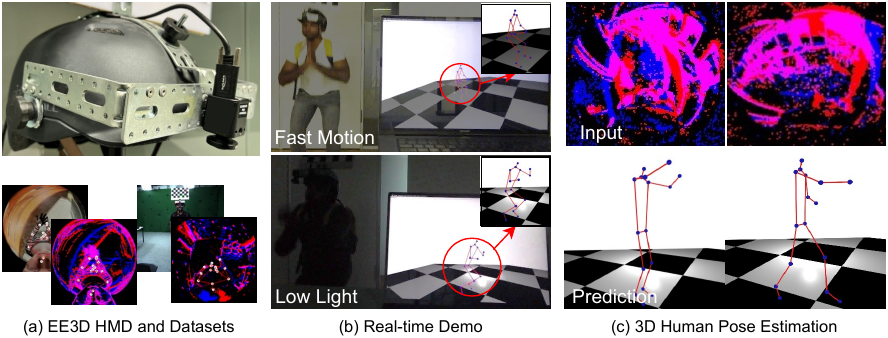}
\end{center}
   \caption{
   \textbf{EventEgo3D++ builds upon the work of EventEgo3D~\citep{Millerdurai_EventEgo3D_2024} for real-time 3D human motion capture from egocentric event streams:}  (a) A photograph of our new head-mounted device (HMD) with a custom-designed egocentric fisheye event camera (top) and visualisations of our synthetically rendered dataset and a real dataset recorded with the HMD (bottom); (b) Real-time demo achieving the pose update rate of $140$Hz; (c) Visualisation of real event streams (top) and the corresponding 3D human poses from a third-person perspective. 
   } 

\label{fig:teaser}
\end{figure*}

Although these experimental prototypes have demonstrated high 3D human pose estimation accuracy, their setups have several limitations. 
Firstly, RGB cameras are prone to over- or under-exposure and motion blur, 
especially in low-light conditions and during rapid movements, which are common in HMD applications.
Secondly, these cameras consume relatively high power, making them less efficient for mobile devices. 
Furthermore, recording image frames synchronously demands high data processing throughput, which can be a significant burden for real-time applications. 
These limitations are particularly problematic for HMDs, where efficient and reliable performance is crucial.

In light of these challenges, our work is motivated by the observation that many of the challenges associated with RGB-based HMDs can be mitigated through the use of \textit{event cameras}.
Event cameras record streams of asynchronous per-pixel brightness changes at high temporal resolution (on the order of microseconds, ${\mu}$s), support an increased dynamic range and consume less power (on the order of tens of $m$W) than RGB cameras, which consume Watts~\citep{gallego2020event}. 
To leverage these benefits, we build a lightweight HMD that integrates an event camera with a fisheye lens. 
This setup allows for the precise capture of fast and dynamic movements with much lower power consumption, making itself well-suited for real-time applications.
Building on these advantages, we develop a lightweight HMD equipped with an event camera and a fisheye lens, enabling precise capture of fast and dynamic movements at notably lower power consumption. 
Further details on event camera efficiency can be found in App.~\ref{appen:event_cam_effi}.

However, existing RGB-based pose estimation techniques, particularly learning-based methods, cannot be straightforwardly repurposed for event streams.
Also, these methods are typically slow and not ideal for real-time applications.
Dedicated approaches are required to fully leverage the advantages of event cameras, as demonstrated by recent progress in event-based 3D reconstruction across various scenarios \citep{EventCap2020, rudnev2021eventhands, Zou2021eventhpe, jiang2023evhandpose, Millerdurai_3DV2024}. 
Furthermore, an egocentric HMD setup utilising an event camera introduces two additional challenges.
Firstly, the \textit{moving event camera} generates a significant amount of background events, making it difficult to isolate the user-specific events required for accurate pose estimation.
Secondly, event cameras fail to generate events in situations where the HMD user remains stationary and no motion is detected.

Our previous work, \EventEgo~\citep{Millerdurai_EventEgo3D_2024} addressed these challenges by introducing a lightweight neural network that processes the egocentric event streams to estimate 3D human pose in real time.
By incorporating confidence scores, the network assigns higher weights to human-generated events than background events, enabling robust pose estimation even in the presence of significant background noise. 
Additionally, a frame buffer mechanism was introduced to maintain stable pose predictions even when only a limited number of events were captured due to the lack of motions.

In this paper, we substantially extend \EventEgo~\citep{Millerdurai_EventEgo3D_2024} with \EventEgoPP, which includes several key improvements and additions. 
Firstly, we improve the 3D pose estimation accuracy of the \EventEgo~framework~\citep{Millerdurai_EventEgo3D_2024} by incorporating additional supervision through a 2D projection loss and a bone loss.
Secondly, in addition to the synthetic dataset (EE3D-S) and the studio-recorded real dataset (EE3D-R) included in \EventEgo, we introduce a new in-the-wild real dataset (EE3D-W) with 3D ground truth poses, providing additional data for fine-tuning and evaluating our method in outdoor environments.
Thirdly, we provide allocentric RGB views and SMPL~\citep{SMPL:2015} body annotations to the real datasets, thereby providing a more comprehensive dataset for advancing research.
The inclusion of in-the-wild data ensures robustness to real-world conditions, while SMPL body annotations provide dense human correspondences, making the datasets valuable for future research and applicable to a wide range of applications.

The remainder of this paper is organised as follows.
Section~\ref{sec:related_work} reviews related work on egocentric 3D human motion capture, event-based 3D reconstruction, and other alternative sensors for 3D human pose estimation.
Section~\ref{sec:method} provides a detailed description of our \EventEgoPP~method, focusing on the neural network architecture and the newly introduced losses.
Section~\ref{sec:setup_datasets} describes the design and implementation of our mobile head-mounted device prototype and the synthetic dataset. Additionally, we outline the recording procedures for the real datasets, including both studio and in-the-wild settings.
Section~\ref{sec:experiments} presents a comprehensive evaluation of our method on synthetic and real datasets. 
Finally, Section~\ref{sec:limitations} discusses the limitations of our approach, and Section~\ref{sec:conclusion} offers our concluding remarks.

\section{Related Work}
\label{sec:related_work} 
We next review related methods for egocentric 3D human pose estimation and event-based 3D reconstruction. 

\subsection{Egocentric 3D Human Pose Estimation}
3D human pose estimation from egocentric monocular or stereo RGB views has been actively studied during the last decade. 
While the earliest approaches were optimisation-based \citep{rhodin2016egocap}, the field promptly adopted neural architectures following the state of the art in human pose estimation. 
Thus, follow-up methods used a two-stream CNN architecture \citep{xu2019mo2cap2} and auto-encoders for monocular \citep{tome2019xr, Tom2023SelfPose3E} and stereo inputs \citep{zhao2021egoglass, hakada2022unrealego,hakada2024unrealego2,Kang2023Ego3DPose}. 
Another work focused on the automatic calibration of fisheye cameras widely used in the egocentric setting \citep{zhang2021automatic}. 
Recent papers leverage human motion priors and temporal constraints for predictions in the global coordinate frame \citep{wang2021estimating}; reinforcement learning for improved physical plausibility of the estimated motions \citep{yuan2019ego, luo2021dynamics}; semi-supervised GAN-based human pose enhancement with external views \citep{wang2022estimating} and depth estimation \citep{wang2023scene}; and scene-conditioned denoising diffusion probabilistic models~\citep{Zhang_2023_ICCV}. 
\citet{Khirodkar_2023_ICCV} address a slightly different setting and use a multi-stream transformer to capture multiple humans in front-facing egocentric views. 
Meanwhile, \citet{wang2024egocentric} focus on egocentric whole-body motion capture with a single fisheye camera, utilising FisheyeViT for feature extraction, specialised networks for hand tracking, and a diffusion-based model for refining motion estimates.

All these works demonstrated promising results and pushed the field forward. 
They, however, were designed for synchronously operating RGB cameras and, hence---as every RGB-based method---suffer from inherent limitations of these sensors (detailed in Sec.~\ref{sec:intro}). 
Thus, only a few of them support real-time frame rates \citep{xu2019mo2cap2, tome2019xr}. 
Moreover, it is unreasonable to expect that RGB-based approaches can be easily adapted for event streams. 
In contrast, we propose an approach 
that (for the first time) accounts for the new data type in the context of egocentric 3D vision (events) and estimates 3D human poses at high 3D pose update rates. 

Last but not least, none of the existing datasets for the training and evaluation of egocentric 3D human pose estimation techniques and related problems \citep{rhodin2016egocap, xu2019mo2cap2, tome2019xr, wang2021estimating, Zhang_ECCV_2022, wang2023scene, Pan_2023_ICCV, Khirodkar_2023_ICCV, wang2022estimating, wang2023egowholebody} provide event streams or frames at framerate sufficient to generate events with event steam simulators \citep{rebecq18a_esim}. 
To evaluate and train our approach, we synthesise and record the necessary datasets (\textit{i.e.}, synthetic, real, and background augmentation) required to investigate event-based 3D human pose estimation on HMDs.

\subsection{Event-based Methods for 3D Reconstruction} 

Substantial discrepancies between RGB frames and asynchronous event data have spurred the development of specialised 3D pose estimation methods, ranging from purely event-based approaches~\citep{rudnev2021eventhands, Nehvi2021, Zou2021eventhpe, Wang2022EvAC3D, xue2022event, chen2022efficient, rudnev2023eventnerf, Millerdurai_3DV2024} to RGB-event hybrid methods~\citep{EventCap2020, Zou2021eventhpe, park20243d, jiang2024complementing}. 
Although hybrid solutions can offer complementary information, they also significantly increase bandwidth usage, power consumption, and computational overhead---factors that become especially problematic for battery-powered head-mounted displays. For a comparison of bandwidth usage and power consumption between RGB and event cameras, please see App.~\ref{appen:event_cam_effi}. 
Consequently, our work adopts a purely event-based paradigm.

Within the event-based domain, \citet{Nehvi2021} track non-rigid 3D objects (polygonal meshes or parametric 3D models) with 
a differentiable event stream simulator. 
\citet{rudnev2021eventhands} synthesise a 
dataset with human hands 
to train a neural 3D hand pose tracker with a Kalman filter. 
They introduce a lightweight LNES representation of events for learning as an improvement upon event frames. 
Next, \citet{xue2022event} optimise the parameters of a 3D hand model by associating events with mesh faces using the expectation-maximisation framework assuming that events are predominantly triggered by hand contours. 
Some works represent events as spatiotemporal points in space and encode them either as point clouds 
\citep{chen2022efficient, Millerdurai_3DV2024}. 
Consequently, most of these approaches are slow (due to different reasons such as iterative optimisation or computationally expensive operations on 3D point clouds), with the notable exception of EventHands \citep{rudnev2021eventhands} 
achieving up to $1$kHz hand pose update rates.

In our work, we leverage LNES \citep{rudnev2021eventhands} because it operates independently of the input event count, facilitates real-time inference, and can be efficiently processed using neural components (\eg~CNN layers). 
Unlike the previously discussed approaches, our method is specifically designed for the egocentric setting and achieves the highest accuracy among all the methods compared.
In particular, we incorporate 
a novel residual %
mechanism that propagates events (event history) from the previous frame to the current one, prioritising events triggered around the human. 
This is also helpful 
when only a few events are triggered due to the lack of motion.

\subsection{Alternate Sensors for 3D Human Pose Estimation} 
Inertial measurement units (IMUs) have been widely used for 3D human pose estimation, often relying on multiple sensors---typically up to six---strategically placed on the head, arms, pelvis, and legs to track body movements~\citep{von2017sparse,huang2018deep,yi2021transpose,jiang2022transformer,yi2022physical}.
While these systems can deliver reasonable accuracy, they tend to be cumbersome and inflexible due to the large number of sensors required and the associated calibration demands. 
Recent advancements have reduced the reliance on multiple sensors, with some systems using as few as three IMUs~\citep{aliakbarian2022flag,winkler2022questsim,jiang2022avatarposer,lee2023questenvsim,jiang2023egoposer,zheng2023realistic,jiang2025manikin}, typically mounted on the head and hands, making them more practical for applications such as virtual reality (VR). 
However, even with fewer sensors, these systems remain prone to issues like sensor drift and frequent recalibration during rapid motion, limiting their effectiveness in high-dynamic scenarios.

Another line of research fuses IMUs with additional modalities such as RGB data~\citep{gilbert2019fusing,von2016human,malleson2017real,guzov2021human,yi2023egolocate,dai2024hmd} or depth maps~\citep{helten2013real}, offering improved global positioning or fine-grained pose estimates. 
Yet, vision-based methods remain sensitive to low-light environments, occlusions, and motion blur, particularly when subjects move rapidly or operate in challenging lighting. 
Although diffusion-based approaches~\citep{du2023agrol,li2023ego,guzov2024hmd} have yielded smoother poses, most rely on future frames to achieve robust predictions, making them unsuitable for real-time usage. 

In contrast, we propose a purely event-camera-based approach, which operates at high frame rates (\ie $140$~fps) and exhibits robustness to challenging conditions like low light and fast motion. 
By mounting a single event camera on a head-mounted display (HMD), we eliminate the need for additional body-worn sensors, thus simplifying the setup and avoiding drift issues. 
This setup not only handles large lighting variations but also naturally accommodates rapid head and body movements, making it especially well-suited for real-time, egocentric 3D human pose estimation.

\begin{figure*}[t]
\begin{center}
   \includegraphics[width=1.0\textwidth]{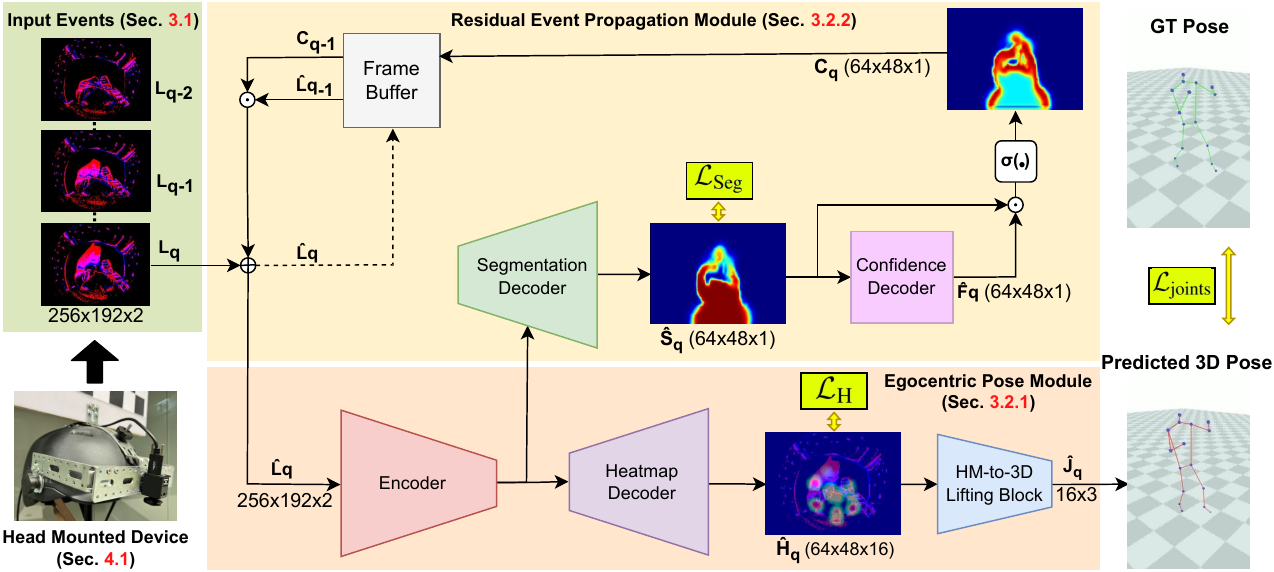}
\end{center}
    \caption{
    \textbf{Overview of our EventEgo3D++ approach}. 
    The HMD captures an egocentric event stream, which is then converted to a series of 2D LNES frames \citep{rudnev2021eventhands} as inputs to our neural architecture to estimate the 3D poses of the HMD user.
    The residual event propagation module (REPM) emphasises events triggered around the human by considering the temporal context of observations (realised with a frame buffer with event decay based on event confidence). 
    REPM, hence, helps the encoder-decoder (from LNES to heatmaps) and the heatmap lifting module to estimate accurate 3D human poses. 
    The method is supervised with ground-truth human body masks, heatmaps and 3D human poses.
}
\label{fig:eventego}
\end{figure*}

\section{The EventEgo3D++ Approach}\label{sec:method} 
Our approach estimates 3D human poses from an egocentric monocular event camera with a fisheye lens.
We first explain the event camera model in Sec.~\ref{subsec:camera_model} and then describe the proposed framework in Sec.~\ref{subsec:EgoEvents3D}.

\subsection{Event Camera Preliminaries}
\label{subsec:camera_model}
Event cameras capture event streams, \ie a 1D temporal sequence that contains discrete packets of asynchronous events that indicate the brightness change of a pixel of the sensor. 
An event is a tuple of the form $e_i = (x_i, y_i, t_i, p_i)$ with the $i$-th index representing the event fired at 
pixel location $(x_i, y_i)$ with its corresponding timestamp $t_i$ and a polarity $p_i \in \{-1, 1\}$. 
The timestamps $t_i$ of modern event cameras have $\mu$s temporal resolution. 
The event 
is generated when the change in logarithmic brightness $\mathbb{L}$ at the pixel location ($x_i, y_i$) exceeds a predefined threshold $C$, 
\textit{i.e.}, $|\mathbb{L}(x_i,y_i,t_i)-\mathbb{L}(x_i,y_i,t_i-t_{p})|\geq{C}$, where $t_p$ represents the
previous triggering time at the same pixel location.
$p = -1$ indicates that the brightness has decreased by $C$; otherwise, it has increased if $p = 1$. 

Modern neural 3D computer vision architectures~\citep{rudnev2021eventhands, lan2023tracking, jiang2023evhandpose} require event streams to be converted to a regular representation, usually in 2D or 3D. 
To this end, we adopt the locally normalised event surfaces (LNES) \citep{rudnev2021eventhands} that aggregate the event tuples into a compact 2D representation as a function of time windows. 
A time window of size $T$ is constructed by collecting all events between the first event $e_0$ (relative to the given time window) and $e_k$, where $t_k - t_0 \leq T$. 
The events from the time window are stored in the 2D LNES frame denoted by $\mathbf{L} \in \mathbb{R}^{H \times W \times 2}$.
For each event within the time window, $e_i \in \{e_1,\hdots, e_k\}$, we update the LNES frame by $L(x_i, y_i, p_i) = \frac{t_i - t_0}{T}$, where an event occurring at pixel location ($x$, $y$) updates the corresponding pixel in the LNES frame.

\noindent \textbf{Note on Visualisation.} For visualisation purposes, we convert each 2-channel LNES frame into a 3-channel (RGB) image by mapping the positive-polarity channel to the red channel, the negative-polarity channel to the blue channel, and setting the green channel to zero.

\subsection{Architecture of EventEgo3D++} 
\label{subsec:EgoEvents3D}
Our approach takes $N$ consecutive LNES frames ${\mathbf{B}=\{\mathbf{L}_1, \hdots, \mathbf{L}_N\}}$, $\mathbf{L}_q{\in} \mathbb{R}^{192 \times 256 \times 2}$ %
as inputs and regresses the camera-centric 3D human body pose per each LNES frame, denoted by 
${\mathbf{O}=\{\mathbf{\hat{J}}_1, \hdots, \mathbf{\hat{J}}_N\}}$,  $\mathbf{\hat{J}}_q{\in}\mathbb{R}^{16 \times 3}$; $q \in \{1, \hdots, N\}$.
$\mathbf{\hat{J}}_q$ include the joints of the head, neck, shoulders, elbows, wrists, hips, knees, ankles, and feet. 
\par
The proposed framework includes two modules; see Fig.~\ref{fig:eventego}. 
First, the Egocentric Pose Module (EPM)
estimates the 3D coordinates of human body joints.
Subsequently, the Residual Event Propagation Module (REPM) propagates events from the previous LNES frame to the current one. 
The REPM module allows the framework 1) to focus more on the events triggered around the human (than those of the background) and 2) to retain the 3D human pose when only a few events are generated due to the absence of motions. 

\begin{figure*}[t]
\begin{center}
   \includegraphics[width=1.0\textwidth]{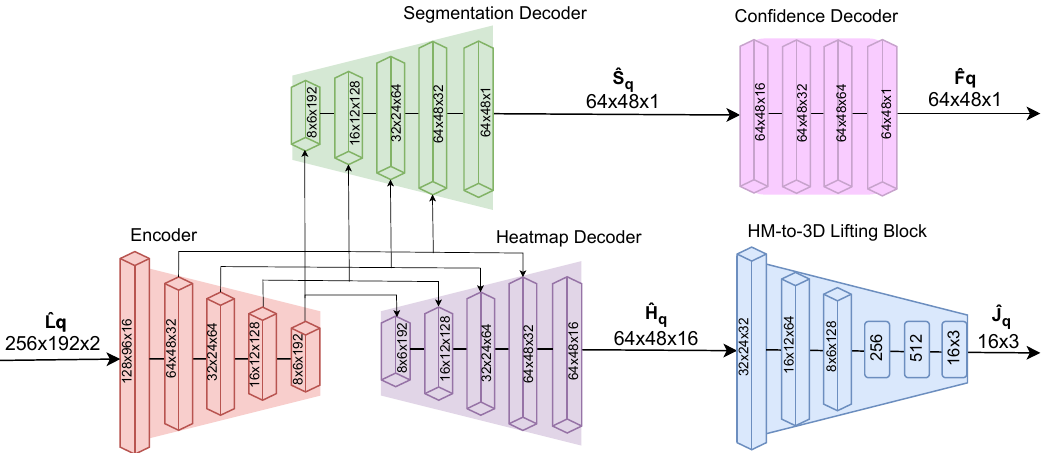}
\end{center}
    \caption{
    \textbf{The network architecture of EventEgo3D++.} The Encoder takes the current LNES frame $\mathbf{\hat{L}}_{q}$ as an input.
    The Heatmap Decoder predicts 2D heatmaps for 16 body joints, which are then fed into the HM-to-3D lifting block to regress 3D joint locations. 
    The Segmentation Decoder generates the human body mask, and the Confidence Decoder subsequently produces a feature map that acts on the human body mask to create a confidence map, highlighting important regions in the egocentric view.
}
\label{fig:network_arch}
\end{figure*}

\subsubsection{Egocentric Pose Module (EPM)}
We regress 3D joints from the input 
$\mathbf{L}_q$ 
in two steps: 1) 2D joint heatmap estimation and 2) the heatmap-to-3D lifting. 
\par
\noindent \textbf{2D joint heatmap estimation}.
To estimate the 2D joint heatmaps, we develop a U-Net-based architecture~\citep{ronneberger2015u}.
Here, we utilise the Blaze blocks~\citep{bazarevsky2020blazepose} as layers of the encoder and decoder to achieve real-time performance. 
The encoder and decoder have five layers each (see Fig. \ref{fig:network_arch}). 
The encoder takes $\mathbf{L}_q$ as input and the heatmap decoder generates 2D joint heatmaps with different resolution sizes from each layer. 
Then, we average them to create the heatmaps of 16 body joints ${\mathbf{\hat H}_q} \in \mathbb{R}^{48 \times 64 \times 16}$ as the final output.
For further details on the heatmap averaging scheme, please refer to App. \ref{appen:2d_hm_est}.

The network is supervised using the mean square error (MSE) between the ground-truth heatmaps and the predicted ones: 
\begin{equation}
\label{eq:losshms}
\mathcal{L}_{\text{H}} = \frac{1}{M_J}\sum_{b=1}^{M_J} \lVert \mathbf{\hat H}_{q,b} \odot V_{q,b} - \mathbf{H}_{q,b} \odot V_{q,b} \rVert ^2, 
\end{equation}
where $\mathbf{\hat H}_{q,b}$ and $\mathbf{H}_{q,b}$ are the predicted and ground-truth heatmaps of the $b$-th joint; \mbox{$V_{q,b} \in \{0, 1\}$} is the visibility of the $b$-th joint; $M_J$ is the number of body joints and $\odot$ is the element-wise multiplication. 
The visibility mask ($V_{q,b}$) ensures that only the joints that are visible and thus relevant for pose estimation contribute to the loss calculation. This is particularly important in scenarios where some joints may be occluded or out of view, such as when the arms are extended or the feet are positioned behind the torso. 
Applying the visibility mask allows the network training to focus more on the joints that are detectable in the input LNES frames instead of occluded or out-of-view joints.

\par
\noindent \textbf{Heatmap-to-3D Lifting Module}.
Following previous works \citep{tome2019xr, pavlakos2018learning}, the Heatmap-to-3D (HM-to-3D) Lifting module takes the estimated heatmaps as input and outputs the 3D joints $\mathbf{\hat J_q} \in \mathbb{R}^{16 \times 3}$. 
This module is based on three convolutional layers and three dense layers (see Fig. \ref{fig:network_arch}).
We supervise the module using three distinct loss terms: 
the MSE of the 3D joints (3D loss), the MSE of the 2D joints reprojected from the 3D joints (2D reprojection loss), and the error in bone orientations and bone lengths (bone loss).

The 3D loss is computed using the ground-truth joint positions and estimated ones at the frame index $q$:
\begin{equation}
\label{eq:lossj3d}
\mathcal{L}_{\text{J3D}} = \frac{1}{M_J}\sum_{r=1}^{M_J} \lVert \mathbf{\hat J}_{q,r} \odot V_{q,r} - \mathbf{J}_{q,r} \odot V_{q,r} \rVert  ^2
, 
\end{equation}
where $M_J$ is the number of body joints, \mbox{$V_{q,r} \in \{0, 1\}$} is the visibility of the $r$-th joint and $\mathbf{\hat J}_{q,r}$ and $\mathbf{J}_{q,r}$ are the predicted and ground-truth $r$-th joint, respectively.

The 2D reprojection loss denoted as $\mathcal{L}_{\text{J2D}}$, compares the 2D projections of the predicted and ground-truth 3D joints, is formulated as:
\begin{equation}
\label{eq:lossj2d}
\mathcal{L}_{\text{J2D}} = \frac{1}{M_J}\sum_{r=1}^{M_J} \lVert 
\Pi(\mathbf{\hat J}_{q,r}) \odot V_{q,r} - \Pi(\mathbf{J}_{q,r}) \odot V_{q,r} \rVert  ^2
, 
\end{equation}
where $\Pi$ is the camera projection function for the fisheye lens, projecting 3D joints into 2D joints.

The bone loss, denoted as $\mathcal{L}_{\text{BA}}$, captures the difference between predicted and ground-truth bone orientations and lengths, allowing the network to learn the spatial relationships between joints and bones. 

For bone orientations, we use a negative cosine similarity loss, defined as:
\begin{equation}
\label{eq:lossja}
\mathcal{L}_{\theta} \;=\; \frac{1}{N_L} \sum_{l=1}^{N_L} \Bigl( 1 \;-\; \cos\bigl(\mathbf{\hat{P}}_{q,l},\, \mathbf{P}_{q,l}\bigr)\Bigr),
\end{equation}
where \(N_L\) is the number of bones, \(\mathbf{\hat{P}}_{q,l} \in \mathbb{R}^3\) is the \(l\)-th predicted bone vector, \(\mathbf{P}_{q,l} \in \mathbb{R}^3\) is the corresponding ground-truth bone vector, and
\[
\cos\bigl(\mathbf{a},\, \mathbf{b}\bigr) \;=\; \frac{\mathbf{a} \cdot \mathbf{b}}{\|\mathbf{a}\|\;\|\mathbf{b}\|}.
\]
This formulation, \(1 - \cos(\cdot,\cdot)\), penalizes misalignment between each predicted bone and its ground-truth counterpart.

For bone lengths, we compute the MSE between the predicted and ground-truth bone vectors, denoted as $\mathcal{L}_{\text{BL}}$:
\begin{equation}
\label{eq:lossbl}
\mathcal{L}_{\text{BL}} = \frac{1}{N_L}\sum_{l=1}^{N_L} \lVert \mathbf{\hat P}_{q,l} - \mathbf{P}_{q,l} \rVert^2.
\end{equation}

The overall bone loss $\mathcal{L}_{\text{BA}}$ is computed by combining the orientation and length losses:
\begin{equation}
\label{eq:lossnl}
\mathcal{L}_{\text{BA}} = 
\lambda_{\theta} \mathcal{L}_{\theta}
+ \lambda_{\text{BL}} \mathcal{L}_{\text{BL}},
\end{equation}
where $\lambda_{\theta}{=}0.001$ and $\lambda_{\text{BL}}{=}0.001$ are the weights assigned to the orientation and length losses, respectively.

Overall, the combined supervision loss for the joints, denoted as  $\mathcal{L}_{\text{joints}}$, is defined as:
\begin{equation}
\label{eq:lossjoints}
\mathcal{L}_{\text{joints}} =
  \lambda_{\text{J3D}} \mathcal{L}_{\text{J3D}}
+ \lambda_{\text{J2D}} \mathcal{L}_{\text{J2D}}
+ \lambda_{\text{BA}} \mathcal{L}_{\text{BA}}
\end{equation}
where we set the weight of each loss as $\lambda_{\text{J3D}}{=}0.01$, $\lambda_{\text{J2D}}{=}0.01$, $\lambda_{\text{BA}}{=}1$.

\subsubsection{Residual Event Propagation Module (REPM)}

In contrast to stationary camera setups, egocentric cameras mounted on head-mounted displays (HMDs) experience diverse movement, which affects the number of events they capture. 
Intense movements by HMD users often result in a large number of events, with a significant portion coming from the background.
Conversely, minimal motion results in very few events.

To address these issues, we introduce the Residual Event Propagation Module (REPM). 
The REPM helps the network focus on events generated by the human body while further incorporating information from previous frames. 
By focusing on human-generated events, the network ensures that these events are given higher importance than background events.
Simultaneously, propagating information from previous frames helps maintain stable pose estimates even when few events are observed.

The REPM comprises the segmentation decoder, the confidence decoder, and the frame buffer.
The segmentation decoder estimates human body masks. 
Next, the confidence decoder takes the body masks as inputs to produce feature maps. 
These feature maps are then used with the body masks to produce confidence maps that indicate regions of the egocentric view to place more importance on.
Lastly, the frame buffer stores the past input frame and its corresponding confidence map, providing weighting to important regions of the current frame (see the top part of Fig.~\ref{fig:eventego}).

\par
\noindent \textbf{Segmentation Decoder}.
The segmentation decoder estimates the human body mask $\mathbf{\hat S}_q \in \mathbb{R}^{48 \times 64 \times 1}$ of the HMD user in the egocentric LNES views.
The architectures of this module and the heatmap decoder are the same except for the final layer that outputs human body masks. 

We use the feature maps from multiple layers of the encoder as inputs to the segmentation decoder (see Fig.~\ref{fig:network_arch}).
The segmentation decoder is supervised by the cross-entropy loss: 
\begin{equation}
\label{eq:loss_seg}
\mathcal{L}_{\text{seg}} = - \mathbf{S}_q \log(\mathbf{\hat S}_q) + (1 - \mathbf{S}_q)\log(1 - \mathbf{\hat S}_q), 
\end{equation}
where $\mathbf{\hat S}_q$ and $\mathbf{S}_q$ are the predicted and ground-truth segmentation masks, respectively. 
\par

\begin{figure}[t]
\begin{center}
   \includegraphics[width=1\columnwidth]{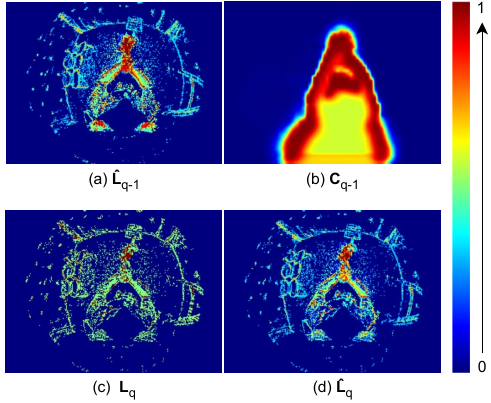}
\end{center} 
    \caption{
        \textbf{Visualisation of frame buffering and human-weighted event generation.}
        The frame buffer holds previous input frame $\mathbf{\hat{L}}_{q-1}$ (a) and previous confident map $\mathbf{C}_{q-1}$ (b). $\mathbf{\hat{L}}_{q-1}$ is weighted with $\mathbf{C}_{q-1}$ and added to the current LNES frame $\mathbf{L}_q$ (c) to produce $\mathbf{\hat L}_{q}$ (d). 
        We can observe that the events generated by the subject are highlighted more than the background events. 
    } 
\label{fig:ed_l_inp}
\end{figure}
\noindent \textbf{Confidence Decoder}.
The confidence decoder is a four-layer convolution network that takes the human body mask $\mathbf{\hat S}_q$ as input and produces a feature map $\mathbf{\hat F}_q \in \mathbb{R}^{48 \times 64 \times 1}$. 
This feature map is then used in combination with $\mathbf{\hat S}_q$ to produce the confidence map $\mathbf{C}_q \in \mathbb{R}^{48 \times 64 \times 1}$: 
\begin{equation}
\label{eq:decay_eqn}
\mathbf{C}_q = \operatorname{sigmoid} (\mathbf{\hat S}_q \odot \mathbf{\hat F}_q),  
\end{equation}
where ``$\operatorname{sigmoid}(\cdot)$'' is a sigmoid operation.

\par
\noindent \textbf{Frame Buffer}. 
The frame buffer stores the previous confidence map $\mathbf{C}_{q-1} \in \mathbb{R}^{48 \times 64 \times 1}$ and the previous input frame $\mathbf{\hat{L}}_{q-1} \in \mathbb{R}^{192 \times 256 \times 2} $. 
Note that we initialise the frame buffer with zeros at the first frame. 
To compute the current input frame $\mathbf{\hat{L}}_{q}$, we retrieve $\mathbf{C}_{q-1} $ and $\mathbf{\hat{L}}_{q-1}$ 
from the frame buffer using the following expression:  
\begin{equation}
\label{eq:ed_output}
\mathbf{\hat{L}}_{q} = (\mathbf{\hat{L}}_{q-1} \odot  \mathbf{C}_{q-1}) \oplus \mathbf{L}_q
\end{equation}
where 
$\mathbf{L}_q$ denotes the LNES frame at the current time and ``$\oplus$'' represents an element-wise addition. 
We normalise the values of $\mathbf{\hat{L}}_{q}$ to the range of $ [-1,1]$. Note, 
$\mathbf{C}_{q-1}$ is resized to $192 \times 256$ before applying Eqn.~\eqref{eq:ed_output}.
See Fig.~\ref{fig:ed_l_inp} for an exemplary visualisation of the components used in Eqn.~\eqref{eq:ed_output}.

\subsubsection{Loss Terms and Supervision} 

Overall, our method is supervised by the heatmap loss $\mathcal{L_{\text{H}}}$ (Eqn.~\ref{eq:losshms}), the joint loss $\mathcal{L_{\text{joints}}}$ (Eqn.~\ref{eq:lossjoints}) and the segmentation loss $\mathcal{L_{\text{seg}}}$  (Eqn.~\ref{eq:loss_seg}) 
as follows:
\begin{equation}
\mathcal{L} = 
  \lambda_{\text{joints}} \mathcal{L}_{\text{joints}} 
+  \lambda_{\text{H}} \mathcal{L}_{\text{H}}
+  \lambda_{\text{seg}} \mathcal{L}_{\text{seg}},
\end{equation}
where we set the weight of each loss as $\lambda_{\text{joints}}{=}1$, $\lambda_{\text{H}}{=}20$, $\lambda_{\text{seg}}{=}0.1$.

\section{Our Egocentric Setup and Datasets}\label{sec:setup_datasets} 
In this work, we introduce three new datasets: EE3D-R, EE3D-W, and EE3D-S. 
These datasets are used to train, evaluate, and fine-tune our \EventEgoPP~method.
EE3D-R and EE3D-W are real-world datasets captured using our head-mounted device (HMD).
The EE3D-R dataset is recorded in a studio environment with controlled lighting and background conditions. 
In contrast, EE3D-W includes both indoor and outdoor environments in the real world, offering a broader range of scenarios that more accurately represent real-world conditions. 
EE3D-S is a large-scale synthetic dataset with the same camera parameters applied from our real-world camera.
EE3D-S provides a diverse array of human poses within a wide variety of virtual backgrounds.
Together, these datasets support a comprehensive approach to developing and refining the \EventEgoPP~method. 
Moreover, pre-training with the synthetic dataset and further fine-tuning on real-world datasets allows the model to handle both diverse and realistic conditions.

\subsection{Real-world Data Capture}
In this section, we first describe our experimental head-mounted device (HMD) used to create real-world datasets, \ie \mbox{EE3D-R} and \mbox{EE3D-W} (Sec.~\ref{sec:hmd}).
Next, we outline the calibration process for our HMD setup (Sec.~\ref{sec:camera_calib}).  
We then detail the procedure for generating ground truth data using the calibrated HMD (Sec.~\ref{subsec:ee3d_gtg}).
Finally, we describe the details of the captured datasets, including their diversity and coverage (Sec.~\ref{subsec:ee3d_real}).

\subsubsection{Head-Mounted Device}
\label{sec:hmd}

\begin{figure}[h]
\center
\includegraphics[width=\linewidth]{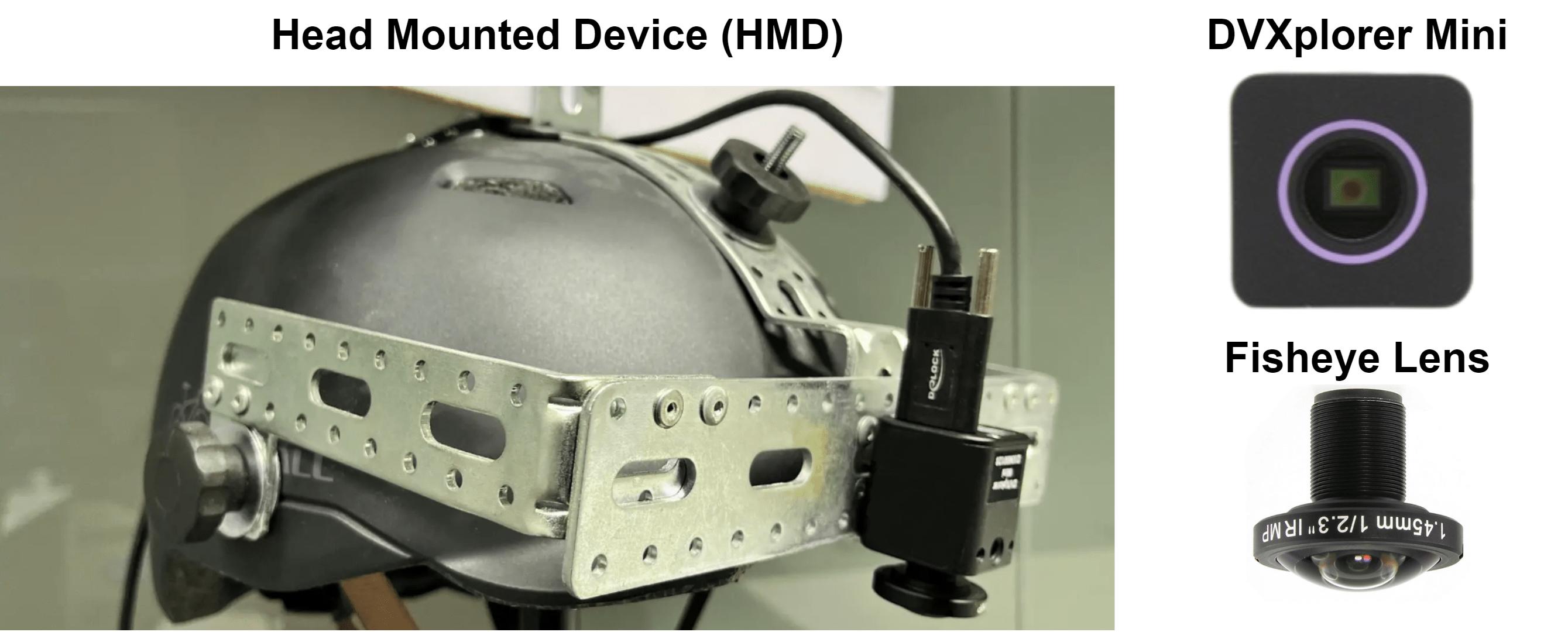}
\caption{\textbf{Our real-world setup.}
The head-mounted device is equipped with an event camera and a fisheye lens.
}
\label{fig:ee3d_hmd}
\end{figure}

Our HMD is a prototypical device consisting of a bicycle helmet with a \citet{DVXplorer_Mini} event camera attached to the helmet $3.5$cm away from the user's head; the strap allows a firm attachment on the head. (see Fig.~\ref{fig:ee3d_hmd})
We use a fisheye lens, \citet{Lensation}, with a field of view of $190\degree$.
The wide field of view effectively covers scenarios where the user's arms are fully extended.
The total weight of the device is ${\approx}0.42$kg. 
The device is used with a laptop in a backpack for external power supply and real-time on-device computing. 
The compact design and the flexibility of our HMD allow users to freely move their heads and perform rapid motions. 

\begin{figure*}[htb]
\center
\includegraphics[width=\linewidth]{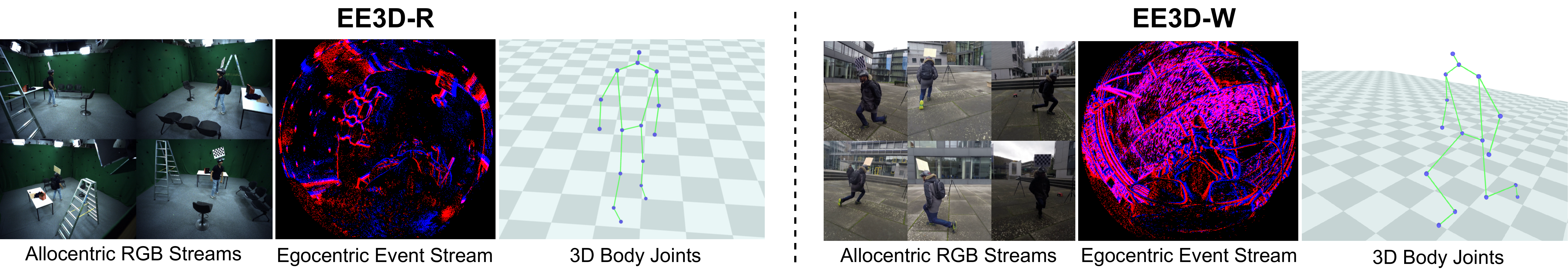}
\caption{
 \textbf{Visualisation of example data from EE3D-R (left) and EE3D-W (right) datasets.} 
}
\label{fig:ee3d_r}
\end{figure*}

\subsubsection{Camera Calibration}
\label{sec:camera_calib}
\noindent \textbf{Intrinsic Calibration.}
We record an event stream of a moving checkerboard, as described by \citet{muglikar2021calibrate}, and then convert the stream into a sequence of images using E2VID~\citep{Rebecq19cvpr}. 
For the intrinsic calibration, we utilise the Scaramuzza projection model~\citep{scaramuzza2006toolbox}, which can account for the radial distortion and the wide field of view of the fisheye lens on our head-mounted device (HMD). 
Specifically, we use MATLAB's Camera Calibrator tool \citep{MATLAB} to obtain the 
projection model parameters.

\noindent \textbf{Extrinsic Calibration.}
To obtain the egocentric 3D poses and SMPL~\citep{SMPL:2015} parameters of the HMD user, we first track the HMD's position during the motion recording. 
This can be achieved by calibrating the HMD equipped with a checkerboard as a reference marker in an allocentric RGB multi-camera setup. 
This step enables us to track the HMD's position within the coordinate frame of the multi-camera setup, \ie the world coordinate frame.
Subsequently, we perform hand-eye calibration to compute the HMD coordinate frame.
Finally, we convert the 3D poses and SMPL parameters from the world coordinate frame into the HMD's coordinate frame.

To obtain the checkerboard images necessary for the hand-eye calibration, we first generate events from the checkerboard and then convert these events into images using the E2VID \citep{Rebecq19cvpr}.
To ensure uniform event distribution, we slide the checkerboard diagonally during the event capture process. 
The final position of the checkerboard after this sliding motion serves as the reference chequerboard position for the calibration procedure.
For additional details on the hand-eye calibration, please refer to App.~\ref{sec:appendix_camera_calib}.

\subsubsection{Ground Truth Generation}
\label{subsec:ee3d_gtg}
We obtain the 3D human poses and SMPL~\citep{SMPL:2015} body parameters
using the multi-view motion capture setups, \citet{captury} and \citet{EasyMocap}.
Specifically, \citet{captury} is a RGB-based multi-view motion capture system that provides accurate human joint positions while \mbox{\citet{EasyMocap}} is used to derive the SMPL parameters from multi-view RGB streams.
Subsequently, we transform these 3D human poses and SMPL parameters from the world coordinate frame to the HMD coordinate frame.
For more details on the accuracy of the generated ground-truth, please refer to App.~\ref{appen:gt_accuracy}.

Additionally, we generate egocentric human body masks, 2D joint coordinates, and joint visibility masks. 
The joint visibility mask indicates whether a joint is visible or occluded from the egocentric view. 
For further details on the generation of the human body masks, 2D egocentric joint coordinates, and joint visibility masks, we refer readers to App.~\ref{sec:appendix_ee3d_gtg}

\noindent \textbf{Note on Additional Metadata.}
Our dataset release includes SMPL body parameters, meshes, and allocentric multi-view RGB streams. 
These supplementary data are provided solely for future research purposes---such as shape estimation and clothing reconstruction---and are not used in the training or evaluation of our framework.

\subsubsection{Real-world Datasets}
\label{subsec:ee3d_real}

Following the procedure in the previous sections, we create the real-world datasets, EE3D-R and EE3D-W.

\noindent \textbf{EE3D-R.} 
EE3D-R is a studio dataset that consists of everyday movements, each performed in different manners by various participants.
We ask twelve subjects---persons with different body shapes and skin tones---to wear our HMD and perform different motions (e.g.~fast) in a multi-view motion capture studio with $30$ allocentric RGB cameras recording at $50$ fps. (see the left part of Fig. \ref{fig:ee3d_r}) 

Each sequence encompasses the following motions: walking, crouching, pushups, boxing, kicking, dancing, interaction with the environment, crawling, sports and jumping.
In the sports category, participants perform specific activities---playing basketball, participating in tug of war, and playing golf. 
Meanwhile, in the interaction with the environment category, the subjects perform actions such as picking up objects from a table, sitting on a chair, and moving the chair.

\begin{figure*}[h]
\centering
   \includegraphics[width=1\linewidth]{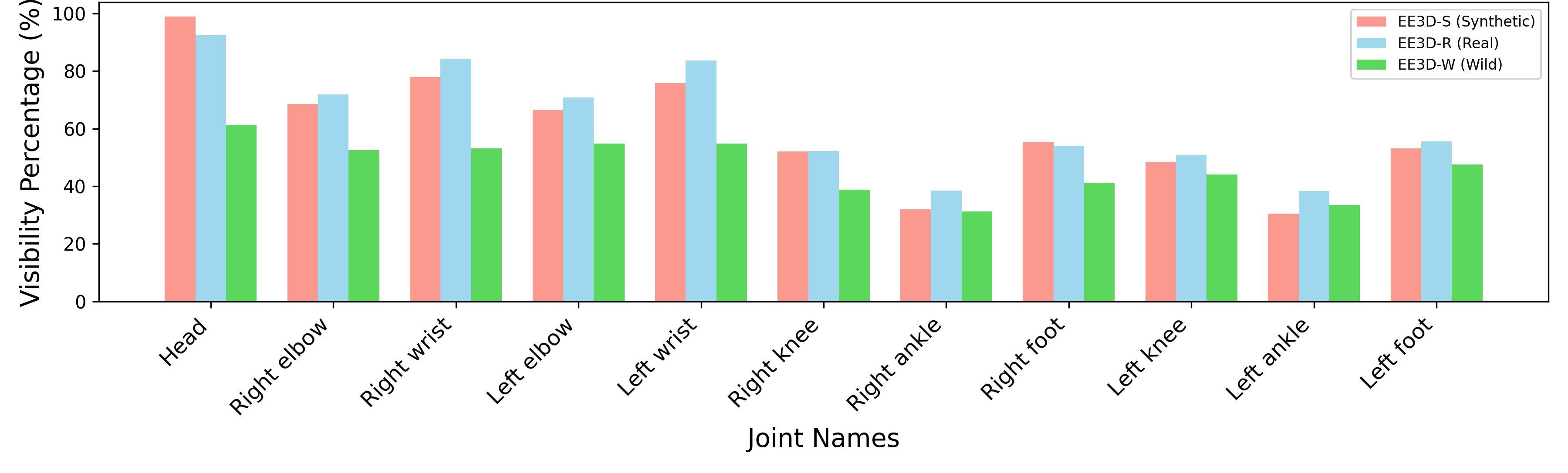}
   \caption{
   \textbf{Joint visibility for SMPL body in our proposed datasets.}
   The visibility percentage is computed as the proportion of samples where each joint is visible from the egocentric perspective.}
\label{fig:ee3d_r_occ_per}
\end{figure*}
In total, we collect 12 sequences containing approximately $4.64 \cdot 10^5$ poses spanning around $155$ minutes. 
These sequences include both fast-paced actions (boxing, kicking, dancing, sports, jumping), comprising approximately \mbox{$2.46 \cdot 10^5$} frames, as well as slower-paced activities in the remaining frames.
Figure~\ref{fig:ee3d_r_occ_per} illustrates the visibility of each joint derived from the SMPL body (see App.~\ref{sec:appendix_ee3d_gtg} for details on the generation process). 
We observe that the lower-body joints are predominantly occluded or out-of-view due to camera constraints, with only about $40\%$ visibility for the ankles.
For our experiments, we use eight sequences ($2.87 \cdot 10^5$ poses) for training, two sequences ($1.05 \cdot 10^5$ poses) for validation, and two sequences ($7.16 \cdot 10^4$ poses) for testing.

\noindent \textbf{EE3D-W.} EE3D-W is an in-the-wild dataset recorded under varying lighting conditions in three different scenes: indoor environments, outdoor areas with concrete flooring, and outdoor areas with grass.
We capture various motions of six subjects in a multi-view motion capture setup with 6 allocentric RGB cameras recording at $60$ fps. (See the right part of Fig. \ref{fig:ee3d_r}.)
The motion types in EE3D-W are similar to those specified in EE3D-R.
This resulted in nine sequences totaling approximately $4.18 \cdot 10^5$ poses over $116$ minutes, with roughly $1.845 \cdot 10^5$ frames containing fast-paced motion. 
As shown in Fig.~\ref{fig:ee3d_r_occ_per}, the in-the-wild dataset exhibits lower overall joint visibility compared to EE3D-R. 
This is primarily because frequent head movements during outdoor activities cause parts of the body to intermittently move in and out of the camera's field of view, thereby increasing occlusions.
For our experiments, we use five sequences ($2.28 \cdot 10^5$ poses) for training, two sequences ($9.32 \cdot 10^4$ poses) for validation, and two sequences ($9.24 \cdot 10^4$ poses) for testing.

\subsection{Synthetic Data Setup} 
\label{sec:ee3d_syn}

In addition to the real-world datasets, we propose EE3D-S, a large-scale synthetic dataset.
In the following, We first describe the virtual human character wearing the HMD and virtual scenes (Sec.~\ref{subsec:EE3DS_Virtual_Human_Character}). 
Next, we explain the rendering and generation of the egocentric event stream (Sec.~\ref{subsec:EE3DS_Rendering}). 
We then outline the ground truth generation for the proposed dataset (Sec.~\ref{subsec:EE3DS_Ground_Truth}). 
Finally, we introduce an event augmentation strategy aimed at reducing the domain gap between real-world datasets (Sec.~\ref{subsec:EE3DS_Event_Augmentation}).

\begin{figure}[h]
\center
\includegraphics[width=\linewidth]{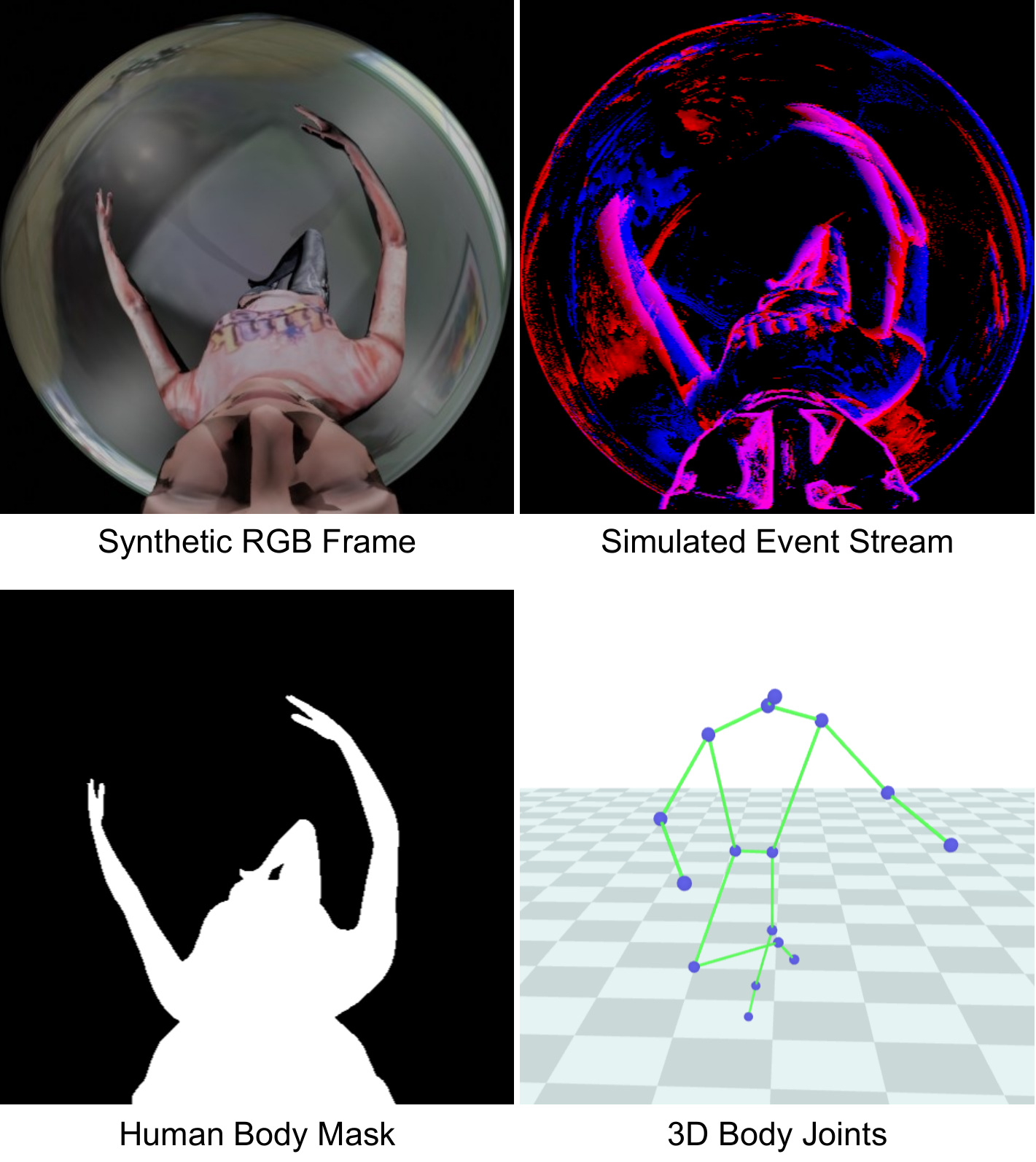}
\caption{
    \textbf{Visualisations of sample data from EE3D-S.} 
}
\label{fig:ee3d_s}
\end{figure}

\subsubsection{Virtual Human Character and Background Scene} 
\label{subsec:EE3DS_Virtual_Human_Character}

We utilise SMPL body models as virtual human users for our HMD, following \citet{xu2019mo2cap2}. 
Body textures are randomly sampled from the SURREAL dataset~\citep{Varol:CVPR:2017}, and animations are driven by motions from the CMU MoCap dataset~\citep{cmumocap}. 
When generating event data, we sample motions at high frame rates~\citep{Gehrig_2020_CVPR} using linear interpolation of SMPL parameters.

As a background scene, we use a $26m^2$ sized 2-dimensional plane with textures sampled from the LSUN dataset~\citep{yu15lsun}.
The scenes are illuminated by four randomly placed point lights within a 5-meter radius of the HMD.

\label{ref:hbm_render}
\subsubsection{Rendering and Event Stream Generation} 
\label{subsec:EE3DS_Rendering}
We render egocentric views using a fisheye camera positioned near the virtual human's face, emulating the real-world HMD setup. 
We apply random perturbations to the fisheye camera position to account for head size variations and HMD movement. 
This allows for simulating real-world scenarios where the camera position relative to the user's head may slightly shift.
We use real-world intrinsic camera parameters (Sec.~\ref{sec:camera_calib}) to render RGB frames and human body masks. 
The rendered RGB frames are then processed by VID2E~\citep{Gehrig_2020_CVPR} to generate the event streams.
Sample data of EE3D-S is shown in Fig.~\ref{fig:ee3d_s}.

In total, we synthesise $946$ motion sequences
containing approximately $6.34 \cdot 10^6$ 3D human poses and $1.419 \cdot 10^{11}$ events.
As shown in Fig.~\ref{fig:ee3d_r_occ_per}, joint visibility is predominantly reduced in the lower body, while the head remains largely unobstructed.
For our experiments, we use 860 sequences ($5.75 \cdot 10^6$ poses) for training, 43 sequences ($3.79 \cdot 10^5$ poses) for validation, and 43 sequences ($2.15 \cdot 10^5$ poses) for testing.
For further details on the configurations used to create the synthetic dataset, we refer readers to App.~\ref{appen:ee3ds_specs}.

\subsubsection{Ground Truth Generation} 
\label{subsec:EE3DS_Ground_Truth}
We extract 3D body joints from the SMPL model, including the head, neck, shoulders, elbows, wrists, hips, knees, ankles, and feet. 
Additionally, we derive 2D joints, human body masks, and visibility masks as outlined in App. \ref{sec:appendix_ee3d_gtg}.

\subsubsection{Event Augmentation}
\label{subsec:EE3DS_Event_Augmentation}

\begin{figure}[h]
\centering
   \includegraphics[width=1\linewidth]{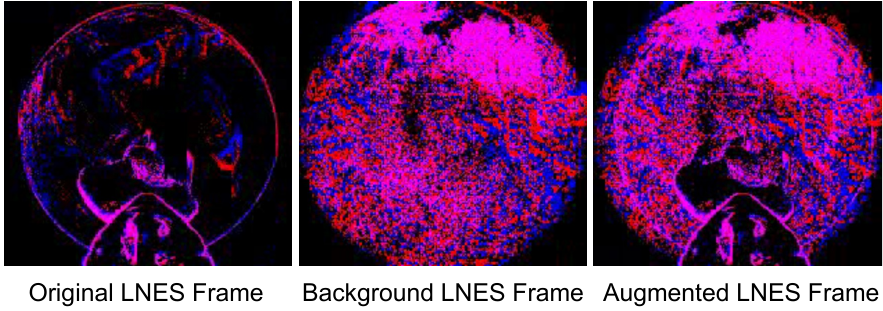}
   \caption{
       \textbf{An example scenario of our event augmentation technique.} The original LNES frame (left) is augmented with an LNES frame of background events (middle) to create an augmented LNES frame (right).
}
\label{fig:dataset_aug}
\end{figure}
Models trained on synthetic data often fail to generalise effectively to real-world scenarios with diverse backgrounds.
To address this issue, we propose an event-wise augmentation technique for background events of the synthetic dataset, EE3D-S (see Fig. \ref{fig:dataset_aug}).
First, we capture sequences of both outdoor and indoor scenes without humans with a handheld event camera, creating background event streams.
These streams are then converted to 2D background LNES frames $\mathbf{L_B}$ (center image in Fig. \ref{fig:dataset_aug}). 
Subsequently, We apply the human body mask \(\mathbf{S_B}\) from EE3D-S to $\mathbf{L_B}$, obtaining a background LNES frame without a region corresponding to a human body in the original LNES frame, denoted as $\mathbf{L_A}$.
Finally, we add \(\mathbf{L_A}\) to the original LNES frames \(\mathbf{L_q}\) from EE3D-S to generate the augmented frame \(\mathbf{L_{\text{aug}}}\) (right image in Fig. \ref{fig:dataset_aug}).
\(\mathbf{L_{\text{aug}}}\) serves as the input to our network.

\section{Experimental Evaluation}\label{sec:experiments}  
This section describes the implementation details of our experiments (Sec.~\ref{ssec:impl}), our results including numerical comparisons to the most related methods (Sec.~\ref{ssec:comparisons}), an ablation study validating the contributions of the core method modules (Sec.~\ref{ssec:ablative}) as well as comparisons in terms of the runtime and architecture parameters (Sec.~\ref{subsection:runtime}). 
Finally, we show %
a real-time demo 
(Sec.~\ref{ssec:qualitative}). 

\subsection{Implementation Details}
\label{ssec:impl}
We implement our method in PyTorch~\citep{paszke2019pytorch} and use Adam optimiser~\citep{Kingma2015} with a batch size of $27$. 
For the EE3D-S dataset, we adopt a learning rate of $10^{-3}$ for $8 \cdot 10^5$ iterations. 
For the EE3D-R dataset, we train our network with a learning rate of $10^{-4}$ for $1.5 \cdot 10^4$ iterations.
For the EE3D-W dataset, we use a learning rate of $10^{-4}$ for $1.2 \cdot 10^4$ iterations.
All modules of our \EventEgoPP~architecture are jointly trained. 
The network is supervised using the most recent ground-truth human pose within the time window $T$ when constructing the LNES frame, \ie the ground-truth pose is aligned with the latest event in the LNES.
We set $T=33$~ms and $N = 20$ for our experiments. 
For additional details on how the LNES frames are constructed, please refer to App.~\ref{appen:input_repres}.
The performance metrics are reported on a single GeForce RTX 3090.
The real-time demo is performed on a laptop equipped with a single 4GB Quadro T1000 GPU housed in a backpack as illustrated in Fig.~\ref{fig:teaser}-(b).

We compare our method \EventEgoPP~with \EventEgo~\citep{Millerdurai_EventEgo3D_2024}, the CVPR version of our work.
In addition, we adapt three existing 3D pose estimation methods for our problem setting: 
\begin{itemize} 
\item 
\citet{xu2019mo2cap2} and \citet{tome2019xr} are egocentric RGB-based methods: We modify their first convolution layer to accept the LNES representation. 
Specifically, we replace the original 3-channel input convolution, which is designed for RGB images, with a 2-channel input convolution layer that is compatible with the LNES representation.

\item 
\citet{rudnev2021eventhands} is an event-based method that takes LNES as input and estimates hand poses: We modify its output layer to regress 3D human poses. 
Specifically, we modify the output linear layer to predict the 3D body joints $\mathbf{\hat{J}} \in \mathbb{R}^{16 \times 3}$.

\end{itemize}
For a fair comparison, we adopt the same training strategy, \ie learning rates and iterations, for all of the competing methods as ours.
We follow previous works \citep{xu2019mo2cap2, zhao2021egoglass, hakada2022unrealego, hakada2024unrealego2, wang2021estimating, wang2022estimating, wang2023scene, wang2023egowholebody} to report the Mean Per Joint Position Error (MPJPE) and MPJPE with Procrustes alignment~\citep{kendall1989survey} (PA-MPJPE).

\subsection{Comparisons to the Related State of the Art}\label{ssec:comparisons} 
\noindent\textbf{Experiment on EE3D-S.}
Firstly, we evaluate our approach on the test set of our synthetic EE3D-S dataset.
To ensure a fair comparison, We train our method and all the competing methods \citep{tome2019xr, xu2019mo2cap2, rudnev2021eventhands, Millerdurai_EventEgo3D_2024} with the training set of our EE3D-S dataset.

From Table~\ref{tab:sotabenchmark_synthetic}, we observe that our method achieves the lowest MPJPE of $98.67$ mm on average, outperforming our previous work~\citep{Millerdurai_EventEgo3D_2024} as well as all other competing methods.
Our method demonstrates superior performance in estimating lower body joints, offering a $9\%$ improvement over \citet{rudnev2021eventhands}, with gains exceeding $11\%$ on the ankle and foot joints. 
This robustness is particularly notable given the significant radial distortion caused by the fisheye lens in our setup, which makes the feet appear much smaller in the input compared to the upper body.
Despite this distortion, our method effectively estimates the position of the feet and other small joint areas, highlighting its accuracy and reliability in challenging conditions.

\begin{table*}[htbp]
\centering
\resizebox{\textwidth}{!}{%
\begin{minipage}{\textwidth}
\renewcommand{\arraystretch}{1.5}
\normalsize
\centering
\resizebox{\textwidth}{!}{%

\begin{tabular}{>{\arraybackslash}m{3.7cm} 
                >{\arraybackslash}m{1.82cm} 
                >{\centering\arraybackslash}m{1.0cm} 
                >{\centering\arraybackslash}m{1.0cm} 
                >{\centering\arraybackslash}m{1.3cm} 
                >{\centering\arraybackslash}m{1.0cm} 
                >{\centering\arraybackslash}m{1.0cm} 
                >{\centering\arraybackslash}m{1.0cm} 
                >{\centering\arraybackslash}m{1.0cm} 
                >{\centering\arraybackslash}m{1.0cm}
                >{\centering\arraybackslash}m{1.0cm}
                >{\centering\arraybackslash}m{1.6cm}}

\toprule
Method & Metric & 
Head &
Neck &
Shoulder &
Elbow &
Wrist &
Hip &
Knee &
Ankle &
Foot  &
Avg. ($\sigma$) \\

\hline

\multirow{2}{*}{\raisebox{-2\height}{\citet{tome2019xr}}} &

MPJPE &
21.33 &
30.80 &
63.07 &
148.09 &
233.09 &
106.88 &
199.07 &
287.17 &
313.75 &
172.15 (97.4) \\&

PA-MPJPE & 
69.93 &
64.59 &
65.75 &
115.83 &
202.93 &
79.62 &
120.17 &
164.88 &
180.53 &
124.62 (49.85) \\
\hdashline

\multirow{2}{*}{\raisebox{-2\height}{\citet{xu2019mo2cap2}}} &

MPJPE &
71.03 &
80.13 &
95.91 &
182.47 &
225.35 &
107.76 &
196.74 &
333.84 &
351.37 &
196.15 (97.98) \\&

PA-MPJPE & 
110.67 &
108.05 &
112.80 &
165.64 &
205.74 &
97.77 &
135.88 &
189.40 &
196.22 &
151.60 (40.48) \\

\hdashline

\multirow{2}{*}{\raisebox{-2\height}{\citet{rudnev2021eventhands}}} &

MPJPE &
\textbf{6.08} &
\textbf{14.11} &
\textbf{31.18} &
76.19 &
99.30 &
\textbf{71.54} &
\textbf{118.20} &
203.14 &
210.92 &
102.57 (68.59) \\&

PA-MPJPE & 
39.00 & 
35.67 &
41.06 &
70.58 &
97.07 &
68.33 &
84.07 &
117.60 &
123.17 &
79.90 (30.07) \\

\hdashline

\multirow{2}{*}{\raisebox{-2\height}{\citet{Millerdurai_EventEgo3D_2024}}} &

MPJPE &
19.41 &
16.38 &
37.23 &
71.43 &
106.61 &
82.97 &
122.88 &
188.19 &
203.20 &
103.80 (62.03) \\&

PA-MPJPE & 

45.60 &
36.05 &
43.09 &
68.22 &
103.91 &
58.89 &
82.55 & 
113.44 &
121.52 &
79.06 (29.47) \\

\hdashline

\multirow{2}{*}{\raisebox{-2\height}{EventEgo3D++ (Ours)}} &

MPJPE &
18.79 &
20.63 &
35.45 &
\textbf{68.24} &
\textbf{97.37} &
73.92 &
118.68 &
\textbf{181.77} &
\textbf{194.26} &
\textbf{98.67} (59.57) \\&

PA-MPJPE & 
\textbf{35.09} &
\textbf{32.13} &
\textbf{36.19} &
\textbf{60.55} &
\textbf{87.17} &
\textbf{51.72} &
\textbf{76.55} & 
\textbf{98.35} &
\textbf{107.00} &  
\textbf{68.89} (26.01) \\

\bottomrule

\end{tabular}%
}
\captionsetup{justification=justified}
\caption{
\textbf{Numerical comparisons on the EE3D-S dataset (in $mm$).}
``$\sigma$'' denotes the standard deviation of MPJPE or PA-MPJPE across body joints.
EventEgo3D++ outperforms all other competing methods, particularly in lower body joints, achieving the best MPJPE. 
Additionally, our method improves lower body performance by  $6\%$ compared to EventEgo3D~\citep{Millerdurai_EventEgo3D_2024}.
}
\label{tab:sotabenchmark_synthetic}
\end{minipage}%
}
\end{table*}

\begin{table*}[htbp]
\centering
\resizebox{\textwidth}{!}{%
\begin{minipage}{\textwidth}
\renewcommand{\arraystretch}{1.5}
\normalsize
\centering
\resizebox{\textwidth}{!}{%

\begin{tabular}{>{\arraybackslash}m{3.8cm} 
                >{\arraybackslash}m{1.82cm} 
                >{\centering\arraybackslash}m{1cm} 
                >{\centering\arraybackslash}m{1cm} 
                >{\centering\arraybackslash}m{1.1cm} 
                >{\centering\arraybackslash}m{1cm} 
                >{\centering\arraybackslash}m{1cm} 
                >{\centering\arraybackslash}m{0.9cm} 
                >{\centering\arraybackslash}m{1.25cm} 
                >{\centering\arraybackslash}m{0.9cm} 
                >{\centering\arraybackslash}m{0.9cm} 
                >{\centering\arraybackslash}m{1cm} 
                >{\centering\arraybackslash}m{1cm} 
                >{\centering\arraybackslash}m{0.9cm} 
                >{\centering\arraybackslash}m{1.9cm}}

\toprule
Method & Metric & 
Walk &
Crouch &
Pushup &
Boxing &
Kick &
Dance &
\makecell{Inter. \\ with env.} &
Crawl &
Sports &
Jump &
Avg.~($\sigma$) \\
 
\hline

\multirow{2}{*}{\raisebox{-2\height}{\citet{tome2019xr}}} &

MPJPE &
140.34 &
173.93 &
157.29 &
177.07 &
181.12 &
212.61 &
169.80 &
144.80 &
207.56 &
165.57 &
173.01 (23.62) \\&

PA-MPJPE & 
104.34 &
119.89 &
102.39 &
124.28 &
121.64 &
132.86 &
111.89 &
88.94 &
120.15 &
110.32 &
113.67 (12.76) \\

\hdashline

\multirow{2}{*}{\raisebox{-2\height}{\citet{xu2019mo2cap2}}} &

MPJPE &
86.09 &
153.53 &
199.34 &
133.15 &
114.00 &
104.44 &
114.52 &
187.95 &
128.21 &
114.10 &
133.53 (36.42) \\&

PA-MPJPE & 
59.11 &
113.31 &
147.13 &
102.50 &
91.75 &
79.65 &
85.83 &
138.12 &
98.10 &
89.19 &
100.47 (26.52) \\
 
\hdashline

\multirow{2}{*}{\raisebox{-2\height}{\citet{rudnev2021eventhands}}} &

MPJPE &
74.82 &
178.23 &
105.68 &
128.93 &
112.45 &
98.14 &
110.05 &
120.51 &
110.16 &
106.19 &
114.52 (26.54) \\&

PA-MPJPE & 
56.77 & 
108.34 &
84.15 &
100.39 &
91.84 &
78.16 &
74.62 &
83.47 &
84.83 &
86.09 &
84.87 (14.08) \\

\hdashline

\multirow{2}{*}{\raisebox{-2\height}{\citet{Millerdurai_EventEgo3D_2024}}} &

MPJPE &
70.88 &
163.84 &
97.88 &
136.57 &
103.72 &
88.87 &
103.19 &
109.71 &
101.02 &
97.32 &
107.30 (25.78) \\&

PA-MPJPE & 

52.11 &
\textbf{99.48} &
75.53 &
104.66 &
86.05 &
71.96 &
70.85 & 
77.94 &
77.82 &
80.17 &
79.66 (14.83) \\

\hdashline

\multirow{2}{*}{\raisebox{-2\height}{EventEgo3D++ (Ours)}} &

MPJPE &
\textbf{68.67} &
\textbf{157.41} &
\textbf{88.63} &
\textbf{123.57} &
\textbf{102.31} &
\textbf{84.95} &
\textbf{95.73} &
\textbf{109.38} &
\textbf{94.9} &
\textbf{95.94} &
\textbf{102.15} (23.01) \\&

PA-MPJPE & 

\textbf{50.06} &
100.76 &
\textbf{66.29} &
\textbf{94.52} &
\textbf{84.26} &
\textbf{66.91} &
\textbf{68.2} & 
\textbf{75.73} &
\textbf{72.23} &
\textbf{75.83} &
\textbf{75.48} (13.95) \\

\bottomrule
\end{tabular}%
}
\captionsetup{justification=justified}
\caption{
\textbf{Numerical comparisons on the EE3D-R dataset (in $mm$).}
``$\sigma$'' denotes the standard deviation of MPJPE or PA-MPJPE across actions.
Our EventEgo3D++ outperforms
existing approaches on most activities by a substantial margin and achieves $11\%$ improvement over \citet{rudnev2021eventhands}.
}
\label{tab:sotabenchmark_full}
\end{minipage}%
}
\end{table*}

\begin{table*}[htbp]
\centering
\resizebox{\textwidth}{!}{%
\begin{minipage}{\textwidth}
\renewcommand{\arraystretch}{1.5}
\normalsize
\centering
\resizebox{\textwidth}{!}{%

\begin{tabular}{>{\arraybackslash}m{3.8cm} 
                >{\arraybackslash}m{1.82cm} 
                >{\centering\arraybackslash}m{1cm} 
                >{\centering\arraybackslash}m{1cm} 
                >{\centering\arraybackslash}m{1.1cm} 
                >{\centering\arraybackslash}m{1cm} 
                >{\centering\arraybackslash}m{1cm} 
                >{\centering\arraybackslash}m{0.9cm} 
                >{\centering\arraybackslash}m{1.25cm} 
                >{\centering\arraybackslash}m{0.9cm} 
                >{\centering\arraybackslash}m{0.9cm} 
                >{\centering\arraybackslash}m{1cm} 
                >{\centering\arraybackslash}m{1cm} 
                >{\centering\arraybackslash}m{0.9cm} 
                >{\centering\arraybackslash}m{1.9cm}}

\toprule
Method & Metric & 
Walk &
Crouch &
Pushup &
Boxing &
Kick &
Dance &
\makecell{Inter. \\ with env.} &
Crawl &
Sports &
Jump &
Avg.~($\sigma$) \\

\hline

\multirow{2}{*}{\raisebox{-2\height}{\citet{tome2019xr}}} &

MPJPE &
469.01 &
555.70 & 
425.97 &
547.19 &
732.93 &
620.50 &
508.09 &
577.70 &
528.96 &
604.00 &
557.01 (81.12) \\&

PA-MPJPE & 
104.11 & 
125.07 &
126.80 &
101.61 &
122.85 &
130.24 &
111.36 &
113.05 &
129.12 &
123.63 &
118.78 (9.90) \\

\hdashline

\multirow{2}{*}{\raisebox{-2\height}{\citet{xu2019mo2cap2}}} &

MPJPE &
218.96 &
234.88 & 
221.28 &
209.71 &
232.84 &
212.79 &
218.56 &
228.13 &
253.04 &
238.96 &
245.32 (12.60) \\&

PA-MPJPE & 
245.41 & 
247.50 &
255.32 &
230.69 &
297.02 &
249.14 &
247.17 &
259.65 &
275.16 &
269.37 &
257.64 (17.79) \\

\hdashline

\multirow{2}{*}{\raisebox{-2\height}{\citet{rudnev2021eventhands}}} &

MPJPE &
\textbf{163.47} &
174.45 &
171.59 &
151.29 &
199.97 &
182.98 &
189.28 &
172.09 &
211.41 &
205.06 &
182.16 (18.23) \\&

PA-MPJPE & 
\textbf{92.29} & 
109.63 &
110.43 &
77.29 &
98.32 &
105.00 &
95.32 &
92.42 &
113.87 &
101.77 &
99.63 (10.38) \\

\hdashline

\multirow{2}{*}{\raisebox{-2\height}{\citet{Millerdurai_EventEgo3D_2024}}} &

MPJPE &
177.70 &
185.86 &
181.70 &
149.22 &
187.12 &
176.62 &
178.65 &
170.90 &
211.38 &
188.90 &
180.81 (14.81) \\&

PA-MPJPE & 

96.77 &
110.64 &
110.62 &
71.12 &
90.05 &
101.32 &
94.23 & 
91.26 &
110.53 &
104.76 &
98.13 (11.74) \\

\hdashline

\multirow{2}{*}{\raisebox{-2\height}{EventEgo3D++ (Ours)}} &

MPJPE &
164.63 &
\textbf{160.88} &
\textbf{171.49} &
\textbf{145.81} &
\textbf{172.32} &
\textbf{163.61} &
\textbf{164.30} &
\textbf{151.32} &
\textbf{193.63} &
\textbf{173.87} &
\textbf{166.19} (12.47) \\&

PA-MPJPE & 

93.44 &
\textbf{96.69} &
\textbf{105.23} &
\textbf{69.62} &
\textbf{89.75} &
\textbf{97.72} &
\textbf{90.33} & 
\textbf{85.12} &
\textbf{104.57} &
\textbf{98.19} &
\textbf{93.07} (9.86) \\

\bottomrule
\end{tabular}%
}
\captionsetup{justification=justified}
\caption{
\textbf{Numerical comparisons on the EE3D-W dataset (in $mm$).} 
``$\sigma$'' denotes the standard deviation of MPJPE or PA-MPJPE across actions.
Our method, EventEgo3D++, outperforms existing approaches with the lowest MPJPE on most activities.
We see an improvement of $10\%$ over \citet{rudnev2021eventhands} in interaction with the environment (Inter.~with env.), showing the robustness of our method against events generated by the environment. 
}
\label{tab:sotabenchmark_wild}
\end{minipage}%
}
\end{table*}

\noindent\textbf{Experiment on EE3D-R.} 
In this experiment, we first pretrain all methods on the EE3D-S dataset. We then fine-tune these methods using the EE3D-R dataset 
and evaluate their performance on the EE3D-R test set.
While the EE3D-S dataset includes a wide range of human motions, there is a domain gap between the synthetic and real-world cases.
This gap arises from factors such as uncontrolled and diverse movement patterns, as well as wearer-specific variability, including differences in posture and movement style.
Fine-tuning the pose estimation methods on real-world data can further reveal their potential in real-world scenarios.

From Table~\ref{tab:sotabenchmark_full}, we observe that our method significantly outperforms all of our comparison methods by a large margin. 
Specifically, our method achieves improvements of $5\%$ in MPJPE on average compared to the best-competing method, \ie \EventEgo~\citep{Millerdurai_EventEgo3D_2024}.
It is also worth noting that our method demonstrates a superiority over the competing methods especially in complex motions involving interaction with the environment, crawling, kicking, sports and dancing.
These motions often come with fast-paced and jittery movements of the HMD, generating substantial background event noise.
Notably, our method excels in handling such challenging scenarios.

\begin{figure}[htbp]
\centering
   \includegraphics[width=1\linewidth]{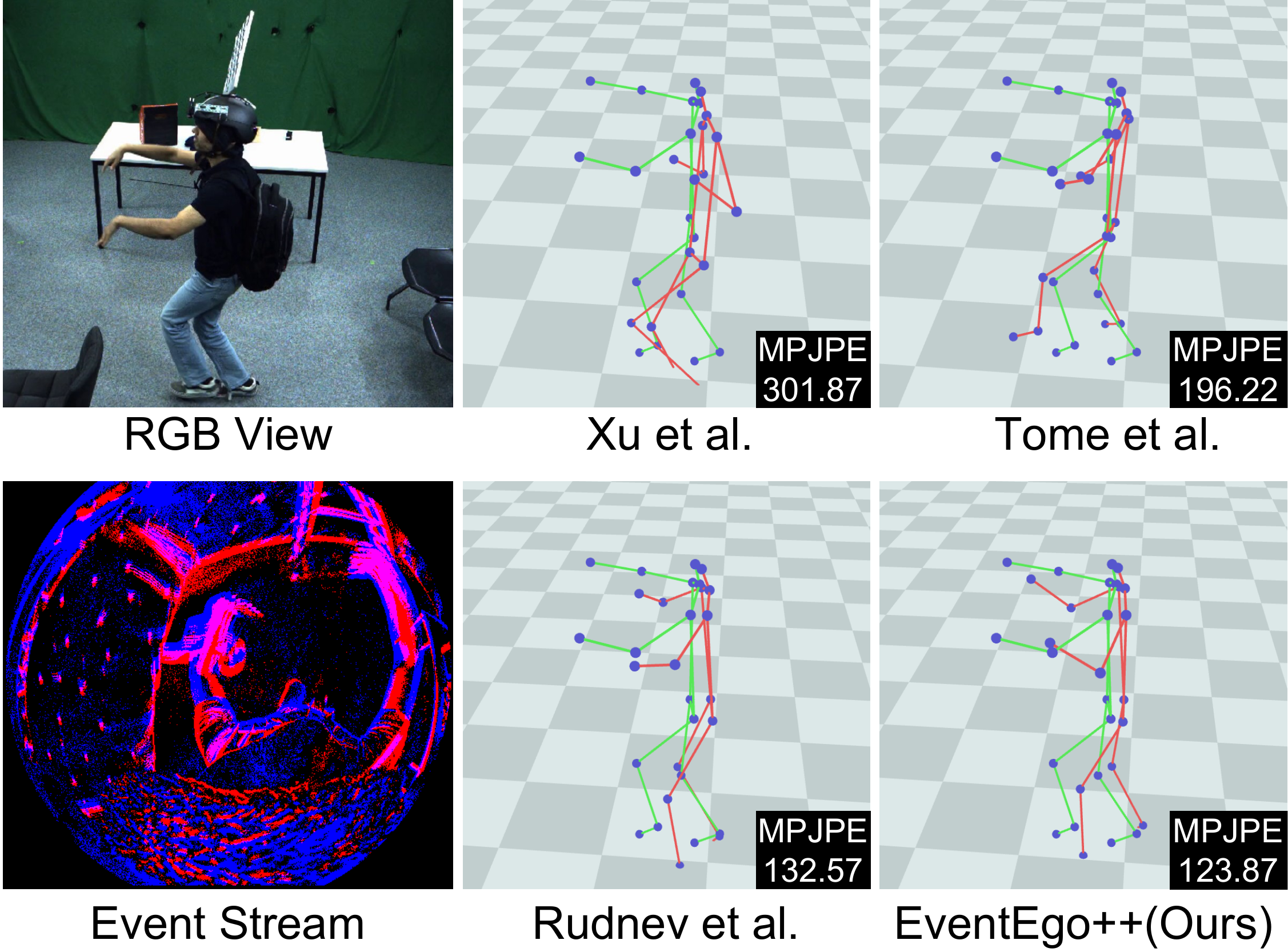}
   \caption{
    \textbf{Qualitative results on EE3D-R}. The MPJPE values are shown in the figures. 3D pose predictions and ground-truth poses are visualised in red and green, respectively.   
    }
\label{fig:qualitative_compare}
\end{figure}

Fig.~\ref{fig:qualitative_compare} shows visual outputs from our approach compared to other methods.
The input LNES frame is noisy and the events generated by the hand sometimes exhibit very close proximity to those generated by the background.
In such scenarios, the competing methods often struggle, predicting incorrect hand positions.
However, our method estimates reasonably accurate 3D poses even in the presence of noisy background events.

\begin{figure}[htbp]
\centering
   \includegraphics[width=1\linewidth]{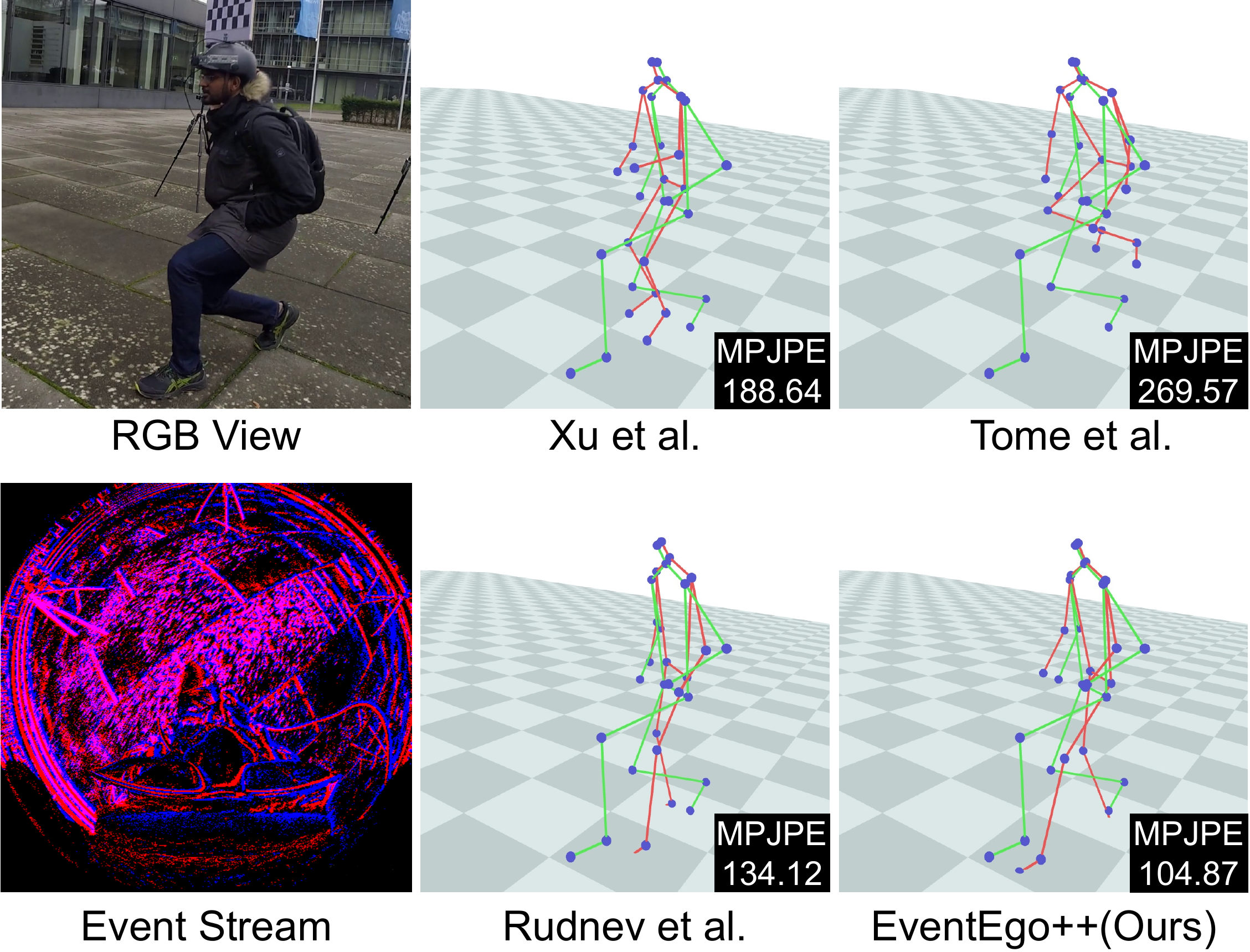}
   \caption{
    \textbf{Qualitative results on EE3D-W}. The MPJPE values are shown in the figures. 3D pose predictions and ground-truth poses are visualised in red and green, respectively.   
    }
\label{fig:qualitative_compare_ee3dw}
\end{figure}

\begin{table}[htbp]
\centering
\resizebox{\columnwidth}{!}{%
\begin{minipage}{\columnwidth}
\renewcommand{\arraystretch}{1.5}
\normalsize
\centering
\resizebox{\columnwidth}{!}{%

\begin{tabular}{lcc}
\hline
Method &
MPJPE &
PA-MPJPE \\
\hline

\multicolumn{1}{l|}{$^\star$\citet{tome2019xr}} &
237.28 &
117.3 \\

\multicolumn{1}{l|}{$^\star$\citet{xu2019mo2cap2}} &
295.17 &
121.91 \\

\hdashline

\multicolumn{1}{l|}{$^\dagger$EventEgo3D++ (Ours)} &
\textbf{102.15} &
\textbf{75.48} \\

\bottomrule

\end{tabular}
}

\caption{
\textbf{Numerical comparisons on the EE3D-R dataset (in $mm$).}
Methods marked with $^\star$ process reconstructed images obtained from event streams using \citet{Rebecq19pami}, while the method marked with $^\dagger$ processes event streams directly. 
}
\label{tab:ee3dr_e2vid} 
\end{minipage}%
}
\end{table}

\begin{figure}[!htbp]
\centering
   \includegraphics[width=1\linewidth]{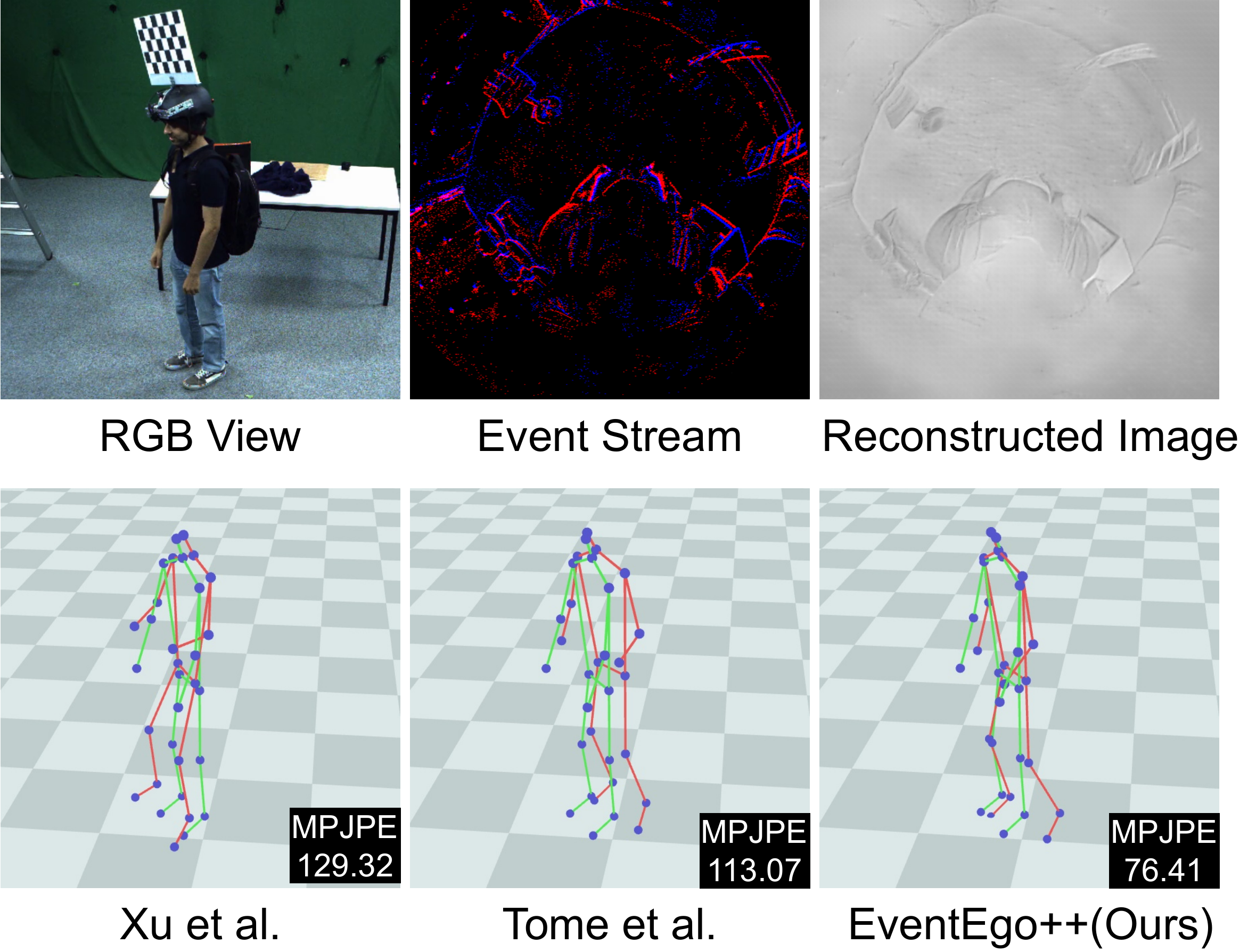}
   \caption{
    \textbf{Qualitative results on EE3D-R}. The MPJPE values are shown in the figures. 3D pose predictions and ground-truth poses are visualised in red and green, respectively.  
    Both \citet{xu2019mo2cap2} and \citet{tome2019xr} process reconstructed images obtained from event streams, whereas EventEgo++ (Ours) directly processes the event streams.
   }
\label{fig:SOTA_E2VID}
\end{figure}

\noindent\textbf{Experiment on image-based reconstructions of EE3D-R.}
In this experiment, we first convert the event streams into image sequences using \citet{Rebecq19pami}. 
We then train and evaluate the RGB-based methods~\citep{xu2019mo2cap2,tome2019xr} on these reconstructed image sequences. 
From Table~\ref{tab:ee3dr_e2vid}, we observe that our method, which directly processes event streams, significantly outperforms the RGB-based methods by a large margin. 
Specifically, we achieve an average improvement of 57\% in MPJPE when compared to the best-performing RGB-based method, \cite{tome2019xr}. 
This performance gap can likely be attributed to artefacts introduced during the image reconstruction process.
When there is significant motion of the person or background, the event camera produces a large number of events, leading to relatively clear reconstructions (see Fig.~\ref{fig:E2VID_examples})
However, in scenarios with sparse events---such as those with slower or minimal motion---the reconstructed images degrade dramatically, making it difficult for RGB-based methods to accurately estimate human poses.
Figure~\ref{fig:SOTA_E2VID} illustrates this issue: although the event data captures the lower body (\eg the right leg), these details are lost in the reconstructed images, leading to poorer performance by RGB-based methods.
In contrast, our method, which leverages the raw event streams, continues to produce reasonably accurate 3D poses even under these challenging conditions.
For additional details on the conversion process, we refer readers to App.~\ref{appen:ee3dr_e2vid}.

\noindent\textbf{Experiment on EE3D-W.} 
We are also interested in pose estimation performance in in-the-wild real-world scenarios, \ie EE3D-W.
Therefore, in this experiment, we initially pretrain all methods on the EE3D-S dataset and then fine-tune them using the training set of EE3D-W for the evaluation on the test set of EE3D-W.
From Tab. \ref{tab:sotabenchmark_wild}, we observe that our approach achieves the best MPJPE and PA-MPJPE scores among all methods.
Compared to other competing methods, there is a significant performance improvement, ranging from a $8\%$ improvement over \citet{rudnev2021eventhands} to a $70\%$ improvement over \citet{tome2019xr} in the MPJPE.
Furthermore, we achieve high accuracy in specific motions, such as crawling, crouching, pushups, and boxing.
This reflects our strength in handling diverse and complex human activities. 
Additionally, we achieve the lowest standard deviation $\sigma$ of the 3D errors on average.
This result indicates that our method is robust across different types of motion, consistently providing accurate 3D pose estimations for a wide range of activities. 
Fig.~\ref{fig:qualitative_compare_ee3dw} shows visual outputs from our approach compared to other methods.
The comparison methods fail to handle the substantial amount of events generated by the background scene.
In this challenging scenario, however, our method estimates reasonably accurate 3D poses.

\subsection{Ablation Study}\label{ssec:ablative} 
\begin{figure}[htbp]
\centering
   \includegraphics[width=1\linewidth]{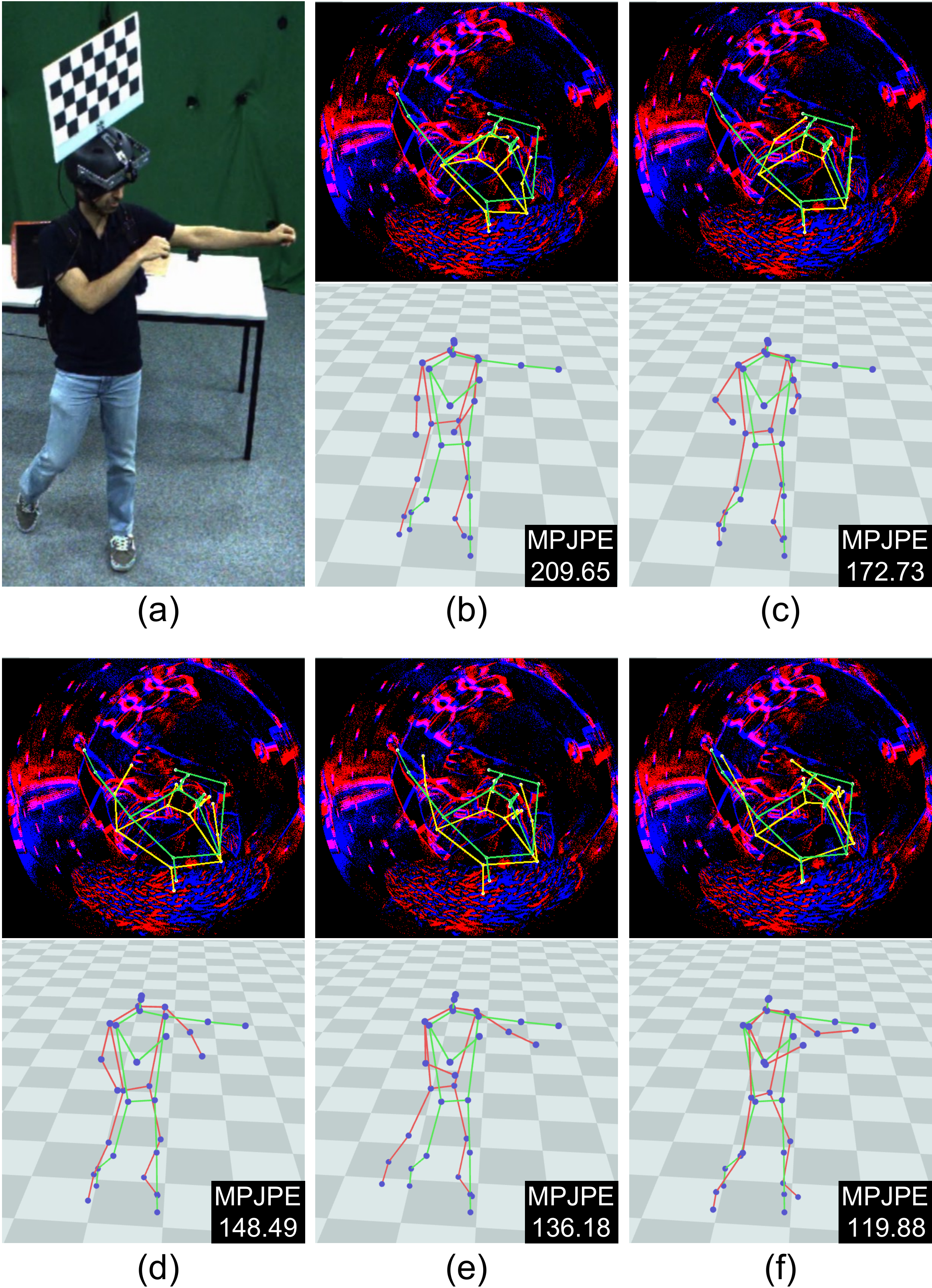}
   \caption{
     \textbf{Qualitative ablation study of our approach on EE3D-R.}  
     (a) Reference RGB view, 
     (b) baseline (EPM only),   
     (c) inclusion of segmentation decoder (Seg. D), 
     (d) inclusion of frame buffer (FB) with Seg. D, 
     (e) inclusion of confidence decoder (Conf. D) with FB and Seg. D, 
     (f) inclusion of 2D reprojection, bone losses, and the visibility mask in (e).
     The MPJPE values are shown in the figures. 3D pose predictions and ground-truth poses are visualised in red and green, respectively. The 2D reprojection of the predicted 3D joints is shown in yellow.  
} 
\label{fig:ablation}
\end{figure}

We next perform an ablation study to systematically evaluate the contributions of the core modules of our method as shown in Tab. \ref{tab:ablation_representation}. 
\begin{table}[htbp]
\centering
\resizebox{\columnwidth}{!}{%
\begin{minipage}{\columnwidth}
\renewcommand{\arraystretch}{1.5}
\normalsize
\centering
\resizebox{\columnwidth}{!}{%

    \begin{tabular}{>{\centering\arraybackslash}m{0.2cm}
                    >{\centering\arraybackslash}m{1.0cm} 
                    >{\centering\arraybackslash}m{0.5cm} 
                    >{\centering\arraybackslash}m{1.2cm} 
                    >{\centering\arraybackslash}m{0.7cm} 
                    >{\centering\arraybackslash}m{0.7cm}
                    >{\centering\arraybackslash}m{0.5cm}
                    >{\centering\arraybackslash}m{1.3cm}
                    >{\centering\arraybackslash}m{1.3cm}}
    \hline
          & Seg. D & FB & Conf. D & $\mathcal{L}_{\text{J2D}}$ & $\mathcal{L}_{\text{BA}}$ & VM & MPJPE & PA-MPJPE \\ \hline
    (I)   &            &            &            &            &            & \multicolumn{1}{c|}{} & 111.01 & 85.58 \\
    (II)  & \checkmark &            &            &            &            & \multicolumn{1}{c|}{} & 108.85 & 84.98 \\
    (III) & \checkmark & \checkmark &            &            &            & \multicolumn{1}{c|}{} & 107.58 & 83.95 \\
    (IV)  & \checkmark & \checkmark & \checkmark &            &            & \multicolumn{1}{c|}{} & 107.30 & 79.66 \\
    (V)   & \checkmark & \checkmark & \checkmark & \checkmark &            & \multicolumn{1}{c|}{} & 106.50 & 77.93 \\
    (VI)  & \checkmark & \checkmark & \checkmark & \checkmark & \checkmark & \multicolumn{1}{c|}{} & 104.73 & 75.79 \\
    (VII) & \checkmark & \checkmark & \checkmark & \checkmark & \checkmark & \multicolumn{1}{c|}{\checkmark} & \textbf{102.15} & \textbf{75.48} \\ \hline
    \end{tabular}%
}
\captionsetup{justification=justified}
\caption{
\textbf{Ablation study of our approach.}
Seg. D (segmentation decoder), FB (frame buffer), Conf. D (confidence decoder), $\mathcal{L}_{\text{J2D}}$ (2D reprojection loss), $\mathcal{L}_{\text{BA}}$ (bone loss) and VM (visibility mask).
We report the MPJPE and PA-MPJPE evaluated on the EE3D-R dataset.
The first row (I) represents the baseline that includes only the egocentric pose module (EPM).
}

\label{tab:ablation_representation} 
\end{minipage}%
}
\end{table}

In Tab.~\ref{tab:ablation_representation}, we first define our baseline method by the Egocentric Pose Module (EPM) without the REPM (I). 
We next systematically examine the impact of the REPM.
Adding the segmentation decoder to the baseline (II) improves the performance by $2\%$ in the MPJPE.
Incorporating the frame buffer along with the segmentation decoder (III) enables past events to propagate to the current frame, resulting in a further $1\%$ improvement in MPJPE.
Additionally, introducing the confidence decoder (IV) significantly enhances performance, \eg by $5\%$ in the PA-MPJPE.
These results validate the effectiveness of each component in REPM.

We also introduce a 2D reprojection loss (V) to refine the alignment of predicted 3D poses with the observed 2D event streams, yielding an additional $1\%$ improvement in MPJPE and a $2\%$ improvement in PA-MPJPE.

The integration of bone loss (VI) and visibility mask  (VII) further improves our method's accuracy.
Specifically, incorporating the bone loss (VI) ensures anatomically plausible bone orientations and lengths, resulting in an additional $1\%$ improvement in MPJPE. 
Furthermore, applying the visibility mask (full model) excludes occluded or out-of-view joints from 3D and 2D joint supervision.
This prevents the model from directly learning the positions of these invisible joints.
Instead, the model estimates their positions based on bone orientations and lengths. 
This approach enables more accurate pose predictions by leveraging the spatial relationships between joints and bones even in cases of occlusion or partial views.
By integrating these losses, our method achieves the best MPJPE and PA-MPJPE scores, with improvements of over $8\%$ and $11\%$, respectively, compared to the baseline. 

\begin{table}[htbp]
\centering
\resizebox{\columnwidth}{!}{%
\begin{minipage}{\columnwidth}
\renewcommand{\arraystretch}{1.5}
\normalsize
\centering
\resizebox{\columnwidth}{!}{%

\begin{tabular}{lccccc}
\hline
Dataset & Config. & $\mathcal{L}_{\text{J2D}}$ & $\mathcal{L}_{\text{BA}}$ & MPJPE & PA-MPJPE \\ 
\hline
\multirow{3}{*}{EE3D-S}  & \multicolumn{1}{|c}{(A)} &                          & \multicolumn{1}{c|}{} & 100.80 & 74.63  \\
                         & \multicolumn{1}{|c}{(B)} & \checkmark               & \multicolumn{1}{c|}{} & 101.27 & 72.76 \\
                         & \multicolumn{1}{|c}{(C)} & \checkmark               & \multicolumn{1}{c|}{\checkmark} & \textbf{98.67}  & \textbf{68.89} \\
\hline
\multirow{3}{*}{EE3D-W}  & \multicolumn{1}{|c}{(A)} &                          & \multicolumn{1}{c|}{} & 177.28 & 100.84\\
                         & \multicolumn{1}{|c}{(B)} & \checkmark               & \multicolumn{1}{c|}{} & 174.30 & 95.76 \\
                         & \multicolumn{1}{|c}{(C)} & \checkmark               & \multicolumn{1}{c|}{\checkmark} & \textbf{166.19} & \textbf{93.07} \\
\hline
\end{tabular}%
}
\captionsetup{justification=justified}
\caption{
\textbf{Ablation study of additional losses.}
$\mathcal{L}_{\text{J2D}}$ (2D reprojection loss) and $\mathcal{L}_{\text{BA}}$ (bone loss). 
We report the MPJPE and PA-MPJPE evaluated on the EE3D-S and EE3D-W datasets with the visibility masks enabled.
}
\label{tab:additional_losses} 
\end{minipage}%
}
\end{table}

To validate these findings across datasets, we evaluate the 2D reprojection and bone loss terms on both EE3D-S and EE3D-W in Tab.~\ref{tab:additional_losses}.
Let (A) represent the model without additional losses.
Adding the 2D reprojection loss (B) consistently reduces errors by a few percentage points on both datasets, indicating that enforcing tight alignment between estimated 3D poses and the 2D projections helps refine pose predictions. 
Furthermore, adding bone loss supervision (C) yields additional improvements in MPJPE, with a larger reduction of 5\% observed in EE3D-W.
This greater improvement is likely due to the more frequent and severe occlusions in the in-the-wild dataset  (see Fig.~\ref{fig:ee3d_r_occ_per}).
By combining bone loss with the other supervisory signals, the model more effectively recovers joint positions by utilising information from nearby visible joints. 
This enables the inference of anatomically consistent poses, even in scenarios where parts of the human body are occluded.

\begin{table}[htbp]
\centering

\renewcommand{\arraystretch}{1.5}
\normalsize

\resizebox{\columnwidth}{!}{%

\begin{tabular}{lcc}
\toprule
Configuration & MPJPE & PA-MPJPE \\ \hline
\multicolumn{1}{l|}{Without Augmentation} & 105.62  & 78.74  \\ 
\multicolumn{1}{l|}{With Augmentation}    & \textbf{102.15}  & \textbf{75.48}  \\
\bottomrule
\end{tabular}
}

\caption{\textbf{Comparison of our approach with and without event augmentation.} Lower values indicate better performance.}
\label{tab:event_aug}

\end{table}

We also examine the impact of event augmentations during pretraining on EE3D-S, as shown in Tab.~\ref{tab:event_aug}. 
Disabling these augmentations degrades generalisation performance on EE3D-R, resulting in a 3\% increase in MPJPE. 
This result highlights the importance of event augmentation in capturing the variability of real-world event noise and preventing the model from overfitting to the training data's limited noise patterns.

\definecolor{printablegreen}{RGB}{34,139,34} 
\newcommand{\greencheck}{\textcolor{printablegreen}{\textbf{\scalebox{1.5}{\checkmark}}}}

\begin{table*}[htbp]
\centering
\renewcommand{\arraystretch}{1.5}
\normalsize
\centering
    \begin{tabular}{>{\centering\arraybackslash}m{0.2cm}
                    >{\centering\arraybackslash}m{1.0cm} 
                    >{\centering\arraybackslash}m{1.0cm} 
                    >{\centering\arraybackslash}m{1.0cm} 
                    >{\centering\arraybackslash}m{1.0cm} 
                    >{\centering\arraybackslash}m{1.0cm}
                    >{\centering\arraybackslash}m{0.7cm}
                    >{\arraybackslash}m{2.3cm}
                    >{\centering\arraybackslash}m{1.3cm}
                    >{\centering\arraybackslash}m{1.7cm}}
\hline
 & $\lambda_{\text{J3D}}$ & $\lambda_\text{H}$ & $\lambda_\text{seg}$ & $\mathcal{L}_{\text{J2D}}$ & $\lambda_\theta$ & $\mathcal{L}_{\text{BL}}$ & Weights & MPJPE & PA-MPJPE \\ 
\hline
\multirow{2}{*}{(II)}  
      & \multirow{2}{*}{\greencheck} & \multirow{2}{*}{} & \multirow{2}{*}{} & \multirow{2}{*}{} & \multirow{2}{*}{} & \multirow{2}{*}{} & 0.01 (current) & 112.29 & 86.39 \\ 
      &                              &                   &                   &                   &                   &                   & 0.1 (10x)      & 112.50 & 86.26 \\ 
\hline
\multirow{2}{*}{(III)} 
      & \multirow{2}{*}{\checkmark} & \multirow{2}{*}{\greencheck} & \multirow{2}{*}{} & \multirow{2}{*}{} & \multirow{2}{*}{} & \multirow{2}{*}{} & 20 (current) & 109.67 & 78.15 \\ 
      &                              &                              &                   &                   &                   &                   & 200 (10x)      & 110.04 & 77.96 \\ 
\hline
\multirow{2}{*}{(IV)} 
      & \multirow{2}{*}{\checkmark} & \multirow{2}{*}{\checkmark} & \multirow{2}{*}{\greencheck} & \multirow{2}{*}{} & \multirow{2}{*}{} & \multirow{2}{*}{} & 0.1 (current) & 108.48 & 78.98 \\ 
      &                              &                              &                              &                   &                   &                   & 1 (10x)      & 108.16 & 77.99 \\ 
\hline
\multirow{2}{*}{(V)} 
      & \multirow{2}{*}{\checkmark} & \multirow{2}{*}{\checkmark} & \multirow{2}{*}{\checkmark} & \multirow{2}{*}{\greencheck} & \multirow{2}{*}{} & \multirow{2}{*}{} & 0.01 (current) & 106.31 & 80.15 \\ 
      &                              &                              &                              &                              &                   &                   & 0.1 (10x)      & 107.20 & 79.99 \\ 
\hline
\multirow{2}{*}{(VI)} 
      & \multirow{2}{*}{\checkmark} & \multirow{2}{*}{\checkmark} & \multirow{2}{*}{\checkmark} & \multirow{2}{*}{\checkmark} & \multirow{2}{*}{\greencheck} & \multirow{2}{*}{} & 0.001 (current) & 102.83 & 76.04 \\ 
      &                              &                              &                              &                              &                              &                   & 0.01 (10x)      & 103.34 & 76.51 \\ 
\hline
\multirow{2}{*}{(VII)} 
      & \multirow{2}{*}{\checkmark} & \multirow{2}{*}{\checkmark} & \multirow{2}{*}{\checkmark} & \multirow{2}{*}{\checkmark} & \multirow{2}{*}{\checkmark} & \multirow{2}{*}{\greencheck} & 0.001 (current) & \textbf{102.15} & \textbf{75.48} \\ 
      &                              &                              &                              &                              &                              &                              & 0.01 (10x)      & 102.50 & 76.15 \\ 
\hline

    \end{tabular}%

\captionsetup{justification=justified}
\caption{
\textbf{Ablation study of loss hyperparamters.}
$\lambda_{\text{J3D}}$ (3D joint loss), $\lambda_\text{H}$ (heatmap loss), $\lambda_\text{seg}$ (segmentation loss), $\lambda_{\text{J2D}}$ (2D reprojection loss), $\lambda_\theta$ (bone orientation loss) and $\lambda_{\text{BL}}$ (bone length Loss). 
\greencheck highlights the loss being ablated, while \checkmark indicates the other losses enabled with their respective "current" weights. 
We report the MPJPE and PA-MPJPE evaluated on the EE3D-R dataset.
}
\label{tab:ablation_loss_hyperparamters} 
\end{table*}

Finally, we present a hyperparameter tuning study in Tab.~\ref{tab:ablation_loss_hyperparamters}, where we vary each loss term's weight by up to a factor of 10. 
Our method exhibits minimal sensitivity to these changes: on average, the MPJPE varies by approximately $1$~mm, suggesting that the contribution of each term remains stable over a broad range of loss weightings.

\begin{figure}[!htbp]
\centering
   \includegraphics[width=1\linewidth]{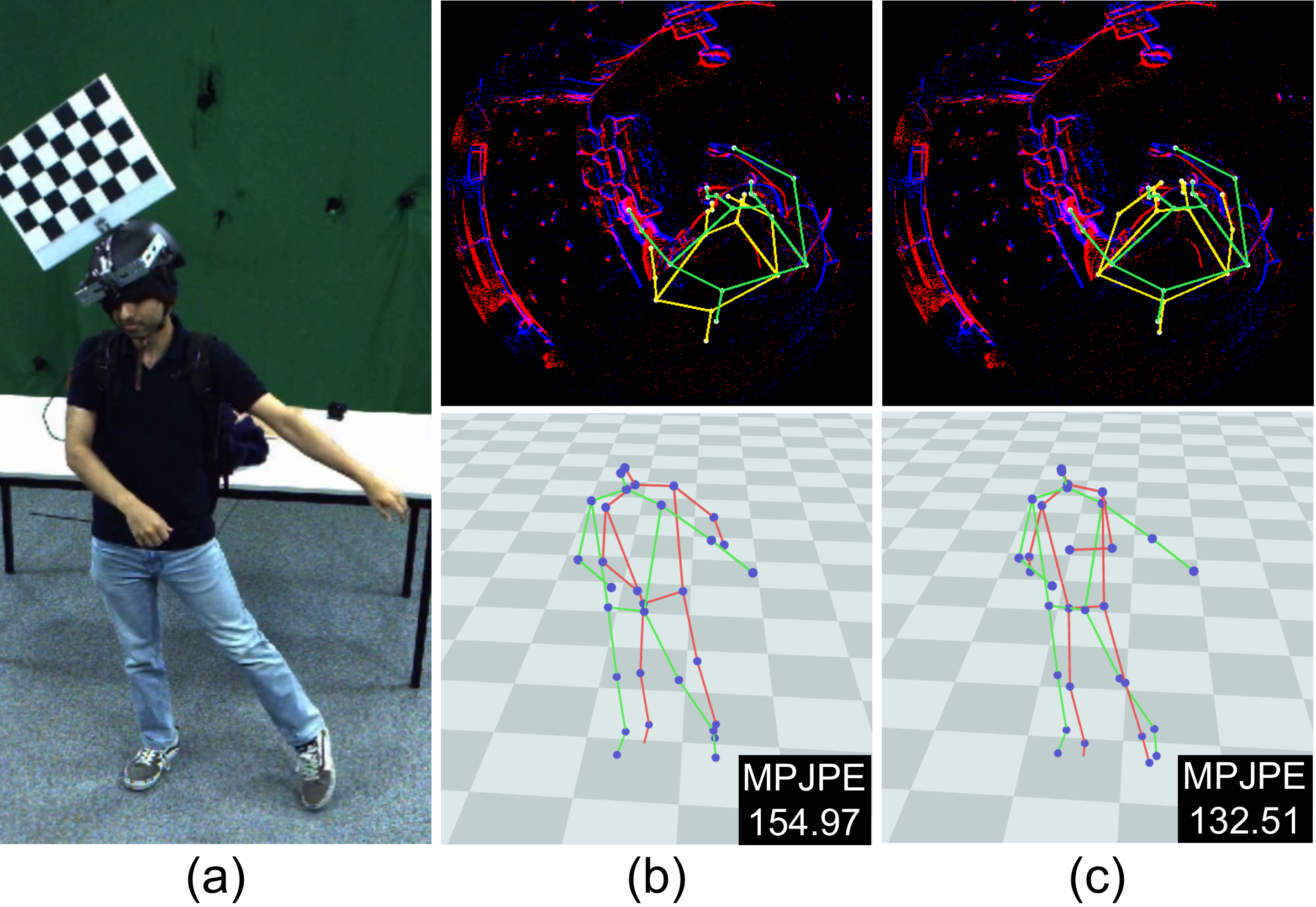}
   \caption{
   \textbf{Qualitative ablation study of 2D reprojection loss on EE3D-R.} (a) Reference RGB view, (b) our model without the loss, (c) inclusion of 2D reprojection loss.
    The MPJPE values are shown in the figures. 3D pose predictions and ground-truth poses are visualised in red and green, respectively. The 2D reprojection of the predicted 3D joints is shown in yellow.     
   }
\label{fig:ablation_reproj_error}
\end{figure}

\begin{figure}[!htbp]
\centering
   \includegraphics[width=1\linewidth]{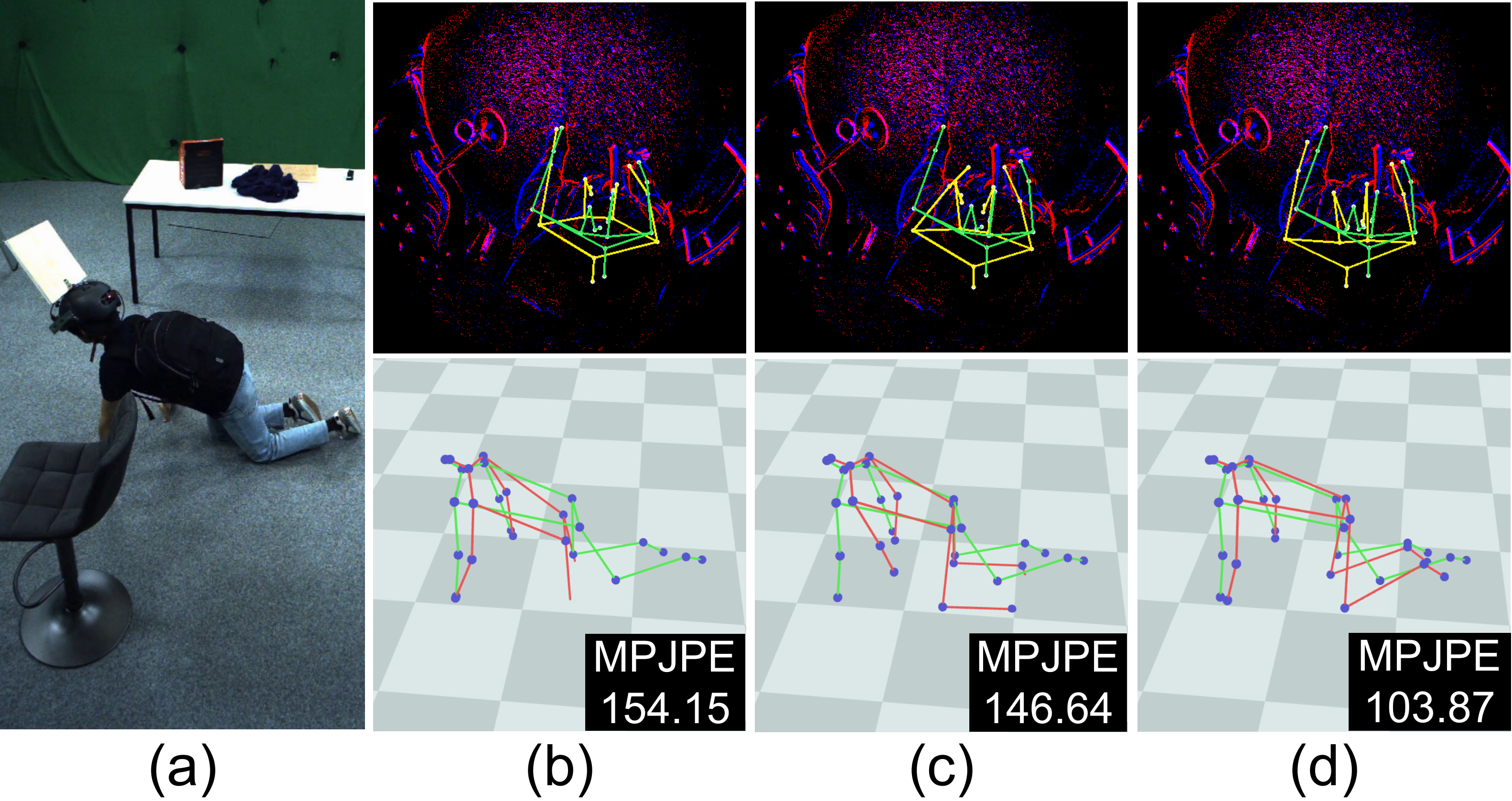}
   \caption{
   \textbf{Qualitative ablation study of bone loss and visibility mask on EE3D-R.} (a) Reference RGB view, (b) our model without bone loss and visibility mask, (c) with bone loss, (d) with both bone loss and visibility mask.
   MPJPE values are displayed. Predicted 3D poses are in red, ground-truth poses are in green, and 2D reprojections are in yellow.
   }
\label{fig:ablation_bone_loss}
\end{figure}

We also provide qualitative ablation studies on the core modules of our approach in Fig.~\ref{fig:ablation}, Fig.~\ref{fig:ablation_reproj_error}, and Fig.~\ref{fig:ablation_bone_loss}. 
From Fig.~\ref{fig:ablation}, we observe that the baseline (b) is highly susceptible to noisy events. 
This significantly affects the network outputs, especially in the hand pose with a very high MPJPE value.
Although this issue can be mitigated by adding the segmentation decoder (c) to some extent, it still struggles to estimate the correct hand position. 
The introduction of Frame Buffer (d) results in a significant performance improvement because it can utilise residual events from the previous frame weighted by the human body mask.
Moreover, the additional inclusion of the confidence decoder (e) further improves the visual quality of pose estimation.
Finally, supervising our framework with the 2D reprojection loss, bone loss and visibility masks (f) plays a key role in producing the best visual outputs.

In Fig.~\ref{fig:ablation_reproj_error}, we visually examine the impact of the 2D reprojection loss (c) in a more challenging motion, such as dancing. 
Similarly, in Fig.~\ref{fig:ablation_bone_loss}, we analyse the influence of bone loss (c) and visibility masks (d) in another demanding motion, namely crawling. 
Despite significant occlusions from the egocentric views, the proposed components enable accurate estimation of human body poses and demonstrate their effectiveness in handling complex scenarios.

\subsection{Runtime Performance} \label{subsection:runtime}

EventEgo3D++ and EventEgo3D~\citep{Millerdurai_EventEgo3D_2024} support real-time 3D human pose update rates of $140$Hz. 
From Tab.~\ref{tab:runtime}, we see that both methods has the lowest number of parameters and floating point operations (FLOPs) compared to the competing methods. 
\citet{rudnev2021eventhands} is the fastest approach and the third-best in terms of 3D accuracy. 
We achieve the second-highest number of pose updates per second. 
This result highlights that our approach is well-suited for mobile devices due to its low memory and computational requirements as well as its low power consumption, due to the event camera.
Since \citet{rudnev2021eventhands} use direct regression of 3D joints, their method is faster, while all other methods use heatmaps as an intermediate representation to estimate the 3D joints.
Furthermore, the operations by \citet{rudnev2021eventhands}~are well parallelisable, which explains its high pose update rate. 
Meanwhile, \citet{xu2019mo2cap2} and \citet{tome2019xr} are not designed for event streams and achieve lower 3D accuracy. 

\begin{table}[htbp]
\centering
\resizebox{\columnwidth}{!}{%
\begin{minipage}{\columnwidth}
\renewcommand{\arraystretch}{1.5}
\normalsize
\centering
\resizebox{\columnwidth}{!}{%

\begin{tabular}{>{\arraybackslash}m{4cm}
                >{\centering\arraybackslash}m{1.2cm} 
                >{\centering\arraybackslash}m{1.2cm} 
                >{\centering\arraybackslash}m{2cm}}

\toprule
 Method & Params & FLOPs & Pose Update Rate \\
 \hline

\text{\citet{tome2019xr}} & 
77.01M      &  11.46G  &  77.07\\
\text{\citet{xu2019mo2cap2}} &  
82.18M & 44.06G & 68.65 \\
\text{\citet{rudnev2021eventhands}} & 
11.2M & 3.58G & \textbf{489.56}\\
\text{\citet{Millerdurai_EventEgo3D_2024}} & 
\textbf{1.25M} & \textbf{416.84M} & 139.88\\
\text{EventEgo3D++ (Ours)} & 
\textbf{1.25M} & \textbf{416.84M} & 139.88\\
\bottomrule
\end{tabular}
}
\captionsetup{justification=justified}
\caption{%
\textbf{Comparisons of model efficiency: number of parameters, FLOPs, and runtime (pose update rate).}
EventEgo3D~\citep{Millerdurai_EventEgo3D_2024} and EventEgo3D++ (Ours) maintain the same number of parameters and FLOPs, achieving the lowest values in both metrics while still maintaining a good pose update rate.
The enhancements in EventEgo3D++ improve accuracy without increasing complexity, refining the EventEgo3D framework.
}
\label{tab:runtime} 
\end{minipage}%
}
\end{table}

\begin{figure*}[htp]
\vspace{-22pt}
\centering
   \includegraphics[width=1\linewidth]{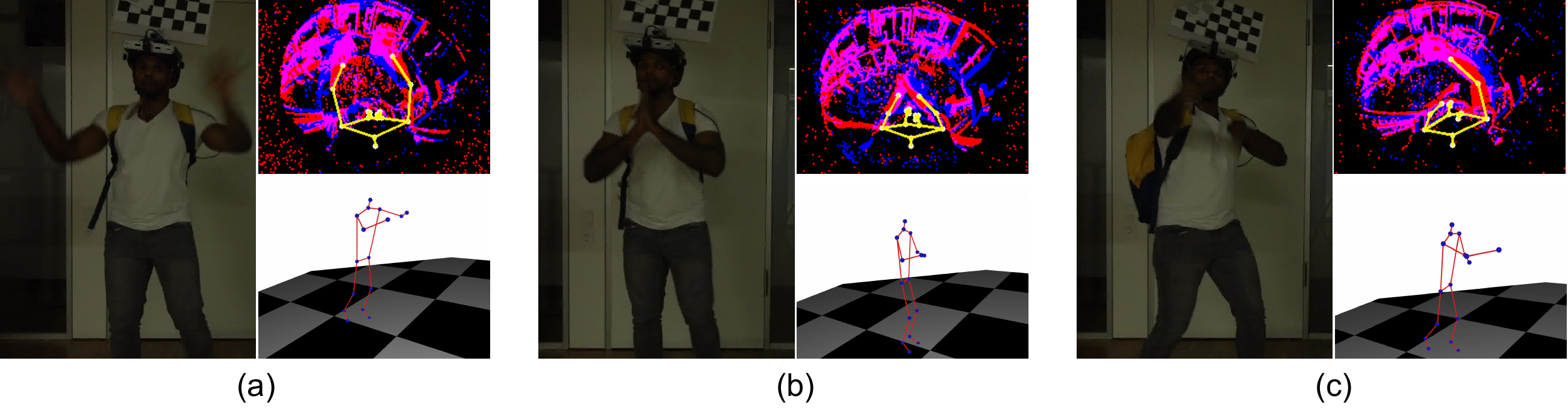}
   \caption{
    \textbf{Qualitative results of our method on in-the-wild motion sequences.} 
         (a) Waving, 
         (b) Clapping and 
         (c) Boxing.
    Our method accurately regresses 3D poses even in low-light conditions. Although the RGB stream experiences significant motion blur due to the fast movement of hands as seen in (a) and (c), our approach effectively utilises the event stream to capture the 3D poses.    
    }
\label{fig:qualitative_results}
\end{figure*}

\subsection{Real-time Demo}
\label{ssec:qualitative} 
Event cameras provide high temporal event resolution and can operate under low-lighting conditions due to their excellent high dynamic range properties. 
\mbox{EventEgo3D++} runs at real-time 3D pose update rates, and we design a real-time demo setup; see Fig.~\ref{fig:teaser}-(b) with a third-person view. 
Our portable HMD enables a wide range of movements, and the on-device computing laptop housed in the backpack allows us to capture in-the-wild sequences. 

We showcase two challenging scenarios, \textit{i.e.}~with fast motions and in a poorly lit environment that would lead to increased exposure time and motion blur in images captured by mainstream RGB cameras. 
Fig.~\ref{fig:qualitative_results} illustrates some of the challenging motions performed during the demo, highlighting that our method accurately estimates 3D poses for each motion. 
Notably, in Fig.~\ref{fig:qualitative_results}-(a), a fast-paced waving motion is depicted, and our method successfully recovers the 3D poses in this dynamic scenario.

\section{Limitations}
\label{sec:limitations}

\EventEgoPP~achieves substantial progress in event-based egocentric pose estimation, particularly in challenging scenarios involving fast motion or low-light conditions, where it surpasses traditional RGB-based methods by producing more robust pose estimates.
Nevertheless, several factors constrain the theoretical "upper bound" of an event-only approach. 
First, event cameras detect changes in brightness rather than absolute intensities. This can cause jitter in the estimated poses when subtle shifts in clothing generate unexpected events, but this is a less pronounced issue in RGB-based methods. 
Second, despite the inherent advantages of event cameras, sensor noise, spurious events, or environmental artefacts (\eg flickering lights) can degrade performance. 
Finally, while our REPM module mitigates the effects of minimal motion by aggregating events, \emph{extended} periods of little or no user movement yield fewer events, allowing sensor noise to dominate and destabilise pose estimates.

Furthermore, our framework employs Locally-Normalised Event Surfaces (LNES; Sec.~\ref{subsec:camera_model}) to convert the event stream into a 2D representation. 
This step may introduce uncertainties when multiple events triggered at the same pixel location within a time window overwrite each other, potentially discarding valuable spatiotemporal details.
Alternative methods, such as those proposed by \cite{chen2022efficient} and \cite{Millerdurai_3DV2024}, aim to preserve the event stream's spatiotemporal representation and could enhance the performance of event-based systems. 
Nonetheless, it is important to note that these methods have been developed for static event cameras. 
When transitioning to \textit{moving event cameras}, new challenges arise, particularly the significant increase in the number of events generated from the background.
While event sampling strategies offer a potential solution to this issue, the effectiveness of importance sampling specifically targeting events generated by the human body remains an unexplored area.
Addressing this challenge could present a promising direction for future research in event-based pose estimation using egocentric cameras.

\section{Conclusion} 
\label{sec:conclusion}
In this work, we present \emph{EventEgo3D++}, an enhanced framework for egocentric 3D human motion capture from event cameras. 
Building upon the existing EventEgo3D framework, EventEgo3D++ introduces additional loss functions and a new in-the-wild dataset (EE3D-W). 
We have further expanded our datasets (EE3D-S, EE3D-R, and EE3D-W) by incorporating parametric human models, as well as allocentric multi-view RGB recordings for the EE3D-R and EE3D-W datasets.
This expanded and diverse dataset provides a comprehensive resource to support and advance future research in the field.
Experimental results demonstrate that EventEgo3D++ achieves state-of-the-art accuracy at real-time pose update rates, excelling in scenarios involving rapid motions and low-light conditions---areas where egocentric event sensing proves particularly advantageous. 
Our method effectively handles sparse and noisy event inputs, maintaining robust performance across a wide range of challenging conditions.
These findings highlight the potential of event-based cameras for egocentric 3D vision tasks and pave the way for future research in areas such as motion analysis, action recognition, and human-computer interaction.

\noindent\textbf{Acknowledgement.}
This research has been partially funded by the ERC Consolidator Grant 4DReply (\textnormal{GA Nr. 770784}) and the EU project FLUENTLY \mbox{(\hspace{-0.1em}\textnormal{GA Nr. 101058680}\hspace{-0.1em})}. Hiroyasu Akada is also supported by the Nakajima Foundation. 

\noindent\textbf{Data Availability.} 
The datasets used in this paper---EE3D-R, EE3D-W, and EE3D-S---are publicly available and can be accessed from the project page at \url{https://eventego3d.mpi-inf.mpg.de}.

\begin{appendices}

\section{Efficiency of Event Cameras}
\label{appen:event_cam_effi}

We evaluate the efficiency of event cameras along two dimensions: (1) the power consumption of our HMD equipped with an event camera, and (2) the bandwidth required to transmit event data over a fixed time window $T$. 

\noindent \textbf{Energy Efficiency of Event Cameras.}
We measure the power draw of the HMD using a precision USB power analyser to record watts (W) and milliamperes (mA). 
On average, the device consumes $\sim\!0.25\,\mathrm{W}$ (\(\sim\!50\,\mathrm{mA}\)), notably lower than typical RGB cameras that often exceed $1~\mathrm{W}$.
Furthermore, no significant variation in power usage is observed between stationary and fast-motion scenarios, whether indoors or outdoors. 
This stability, despite rapid head movements or dynamic backgrounds, highlights the suitability of event cameras for continuous, real-time egocentric applications.

\begin{figure*}[!htbp]
\centering
   \includegraphics[width=1\linewidth]{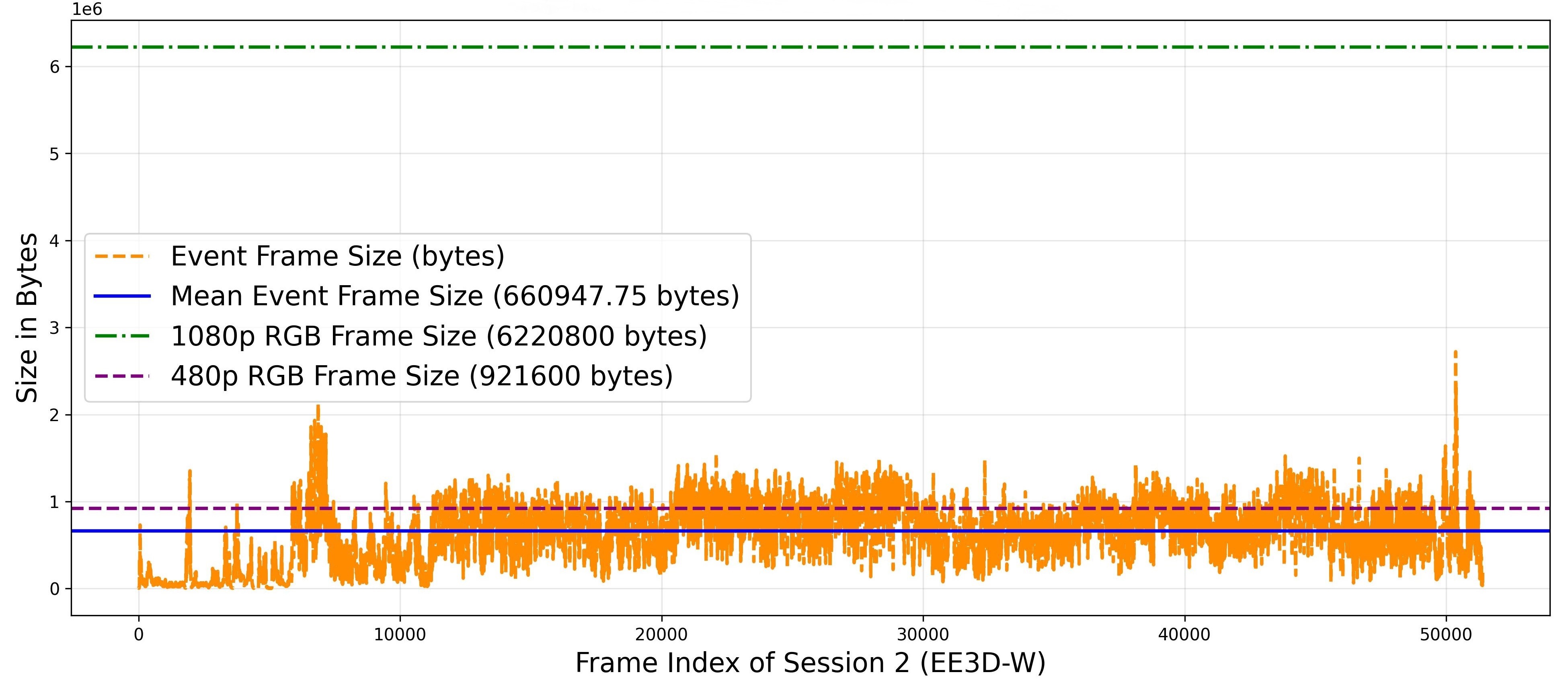}
   \caption{
    \textbf{Bandwidth comparison between event streams and RGB frames.}
   }
\label{fig:bandwidth}
\end{figure*}

\noindent \textbf{Event Camera Bandwidth Requirements.}
We measure the bandwidth consumption on a representative EE3D-W sequence (\emph{S2}), featuring outdoor, in-the-wild conditions that generate a large number of events from both the wearer's body and the background.
Fig.~\ref{fig:bandwidth} plots the per-frame bandwidth usage for this sequence, showing an average of approximately $6.6 \cdot 10^{5}$ bytes per frame. 
Each event is a $13$-byte tuple $(x, y, t_s, p)$, where $\mathit{x}$ and $\mathit{y}$ each require $4$ bytes, $\mathit{t_s}$ requires $8$ bytes, and $\mathit{p}$ requires $1$ byte.  
These events are accumulated over a time window $T=16.66~\mathrm{ms}$, matching the $60~\mathrm{fps}$ rate of the allocentric RGB cameras.
By comparison, an RGB frame at \mbox{$1920\times1080$} encodes each pixel in 3 bytes (RGB), resulting in \mbox{$1920\times1080\times3 \approx 6.22 \cdot 10^{6}$} bytes per frame---about $9.4$ times higher than our event-stream data.
Even at a lower resolution of $640\times480$, which matches our event camera, RGB data requires about $1.39$ times more bandwidth than the event stream.

\section{2D Joint Heatmap Estimation}
\label{appen:2d_hm_est}

We estimate 2D joint heatmaps using the Heatmap Decoder. 
We produce heatmaps at different resolutions from the layers of the decoder.
Specifically, we utilise layers $2$, $3$, $4$, and $5$, extracting the first $16$ feature maps from each layer. 
Each feature map corresponds to a heatmap for each body joint.
These heatmaps are then upsampled to a common resolution of
$48 \times 64$.
After upsampling, we average the heatmaps from all the selected layers to produce the final heatmaps ${\mathbf{\hat H}_q} \in \mathbb{R}^{48 \times 64 \times 16}$, which represent the 2D joint heatmaps for the body joints.

\section{Real World Data Capture}

\subsection{Head Mounted Device Calibration}
\label{sec:appendix_camera_calib}

To obtain the ground-truth pose of the HMD user, we first calibrate the HMD using an allocentric RGB multi-camera setup. 
This calibration allows us to determine the HMD's position in the 
multi-camera setup's coordinate frame \ie the world coordinate frame.
Finally, we compute the world-to-device transformation matrix, denoted by $\mathbf{M}_{\text{WE}}$, which maps the world coordinate frame to the HMD coordinate frame. 
This lets us obtain the user's 3D pose within the HMD's coordinate system.

The position of the HMD in the world coordinate frame is obtained through hand-eye calibration, following the approach of \citet{rhodin2016egocap}. 
In this process, a checkerboard, referred to as the "head-checkerboard," is mounted on top of the HMD. 
This checkerboard is a surrogate for the event camera's position, enabling precise tracking of the HMD within the world coordinate system.
We compute the $\mathbf{M}_{\text{WE}}$ matrix in two steps. First, we obtain the transformation from the world to the head-checkerboard coordinate frame, denoted as $\mathbf{M}_{\text{WC}}$.
Next, we calculate the transformation from the head-checkerboard to the event camera, denoted by
$\mathbf{M}_{\text{CE}}$.
Specifically, $\mathbf{M}_{\text{WE}}$, is defined as:
\begin{equation}
\label{eq:WE}
\mathbf{M}_{\text{WE}} =  ({{\mathbf{M}_{\text{CE}}} \cdot {\mathbf{M}_{\text{WC}}}})
\end{equation}

\begin{figure}[h]
\center
\includegraphics[width=\linewidth]{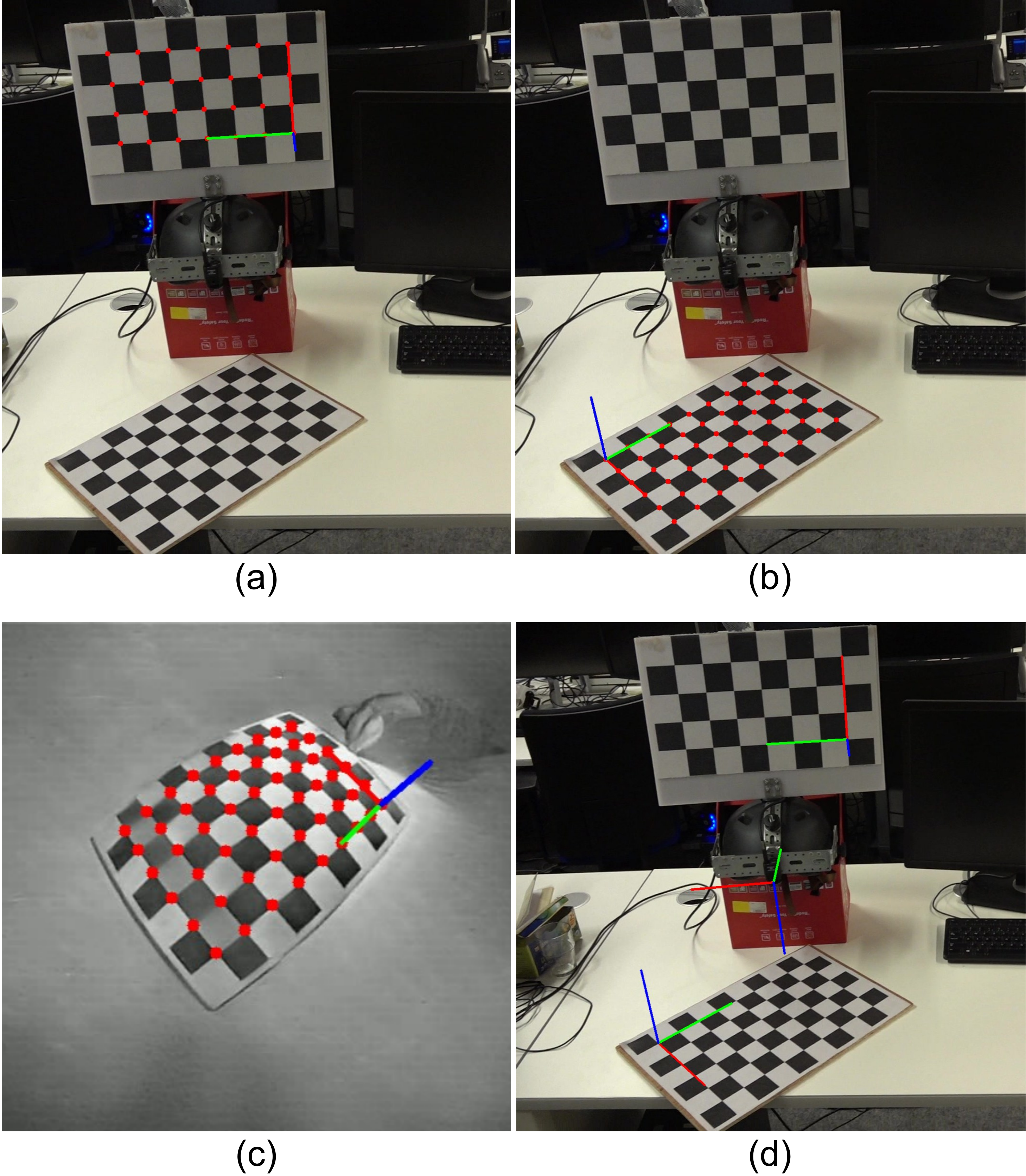}
\caption{
\textbf{Hand-eye calibration for determining event camera position relative to checkerboard on head-mounted device}
(a) The coordinate frame of the head-checkerboard is obtained using the external RGB camera.
(b) The coordinate frame of the floor-checkerboard is obtained using the external RGB camera.
(c) The coordinate frame of the floor-checkerboard is obtained using the event camera.
(d) After hand-eye calibration is performed, the event camera
is localised with respect to the head-checkerboard. 
}
\label{fig:fisheye_calib}
\end{figure}

The $\mathbf{M}_{\text{WC}}$ matrix is obtained by solving the pose of the head-checkerboard in the world coordinate frame. 
We apply the PnP algorithm~\citep{itseez2015opencv} on the images obtained from the multi-view RGB setup for the pose computation.
Meanwhile, the $\mathbf{M}_{\text{CE}}$ matrix is obtained through the following steps:
\begin{itemize}
\item 
Generate a checkerboard image using the event camera: we first capture an event stream of a checkerboard placed at the bottom HMD, referred to as the "floor-checkerboard," while keeping the HMD stationary. 
To create a uniform distribution of events in both vertical and horizontal directions, the checkerboard is slid diagonally. 
The captured event stream is then converted into image sequences using E2VID \citep{Rebecq19cvpr}. 
From these sequences, we select the image that captures the last position of the floor-checkerboard after the slide. 
Finally, we compute its pose, $\mathbf{M}_{\text{E}}$, in the HMD coordinate system using the PnP algorithm.
A visualisation is shown in Fig. \ref{fig:fisheye_calib}-(c).
\item 
While maintaining the positions of both the floor-checkerboard and the HMD from the previous step, we use an external RGB camera to capture an image sequence that includes both the head-checkerboard and floor-checkerboard.
We then select the images where the calibration patterns for both checkerboards are detected. 
With these selected images, we compute the poses of the head-checkerboard ($\mathbf{M}_{\text{H}}$) and floor-checkerboard ($\mathbf{M}_{\text{F}}$) relative to the external RGB camera using the PnP algorithm.
\item
Finally, the $\mathbf{M}_{\text{CE}}$ matrix is obtained through the following transformation:
\begin{equation}
    \label{eq:CE}
    \mathbf{M}_{\text{CE}} = \mathbf{M}_{\text{E}} \cdot {\mathbf{M}_{\text{F}}}^{-1} \cdot \mathbf{M}_{\text{H}}. 
\end{equation}
A visualisation of the calibrated setup is shown in Fig. \ref{fig:fisheye_calib}-(d).
\end{itemize}

\subsection{Accuracy of Ground Truth}
\label{appen:gt_accuracy}
We acquire 3D human poses and SMPL~\citep{SMPL:2015} parameters using two multi-view motion capture pipelines: \citet{captury} for accurate 3D joints and \citet{EasyMocap} for SMPL parameter recovery.

\noindent \textbf{EE3D-R Dataset.} Captured with a state-of-the-art commercial system~\citep{captury} at $50$~fps under high illumination, EE3D-R uses $30$ cameras to minimise motion blur and maximise tracking accuracy. 
This setup aligns with prior literature on multi-view pose capture~\citep{EventCap2020,wang2021estimating,wang2022estimating,wang2023scene,wang2023egowholebody,hakada2024unrealego2,wang2024egocentric,Millerdurai_3DV2024} and ensures robust 3D reference poses.

\noindent \textbf{EE3D-W Dataset.} In contrast, EE3D-W is filmed at $60$~fps using $6$ cameras in outdoor settings, leveraging the same \citet{captury} technology. 
Although fewer cameras are employed, the system remains sufficient for accurate ground-truth capture, following best practices used in prior works for outdoor environments~\citep{elhayek2016marconi,singleshotmultiperson2018,xu2019mo2cap2}. 

In both datasets, each event in the egocentric event stream is synchronised with the allocentric RGB frames up to the frame's timestamp. 
Together, EE3D-R and EE3D-W provide diverse, well-calibrated benchmarks, facilitating robust evaluations of egocentric 3D human pose estimation.

\subsection{Ground Truth Generation}
\label{sec:appendix_ee3d_gtg}
\begin{figure}[h]
\center
\includegraphics[width=\linewidth]{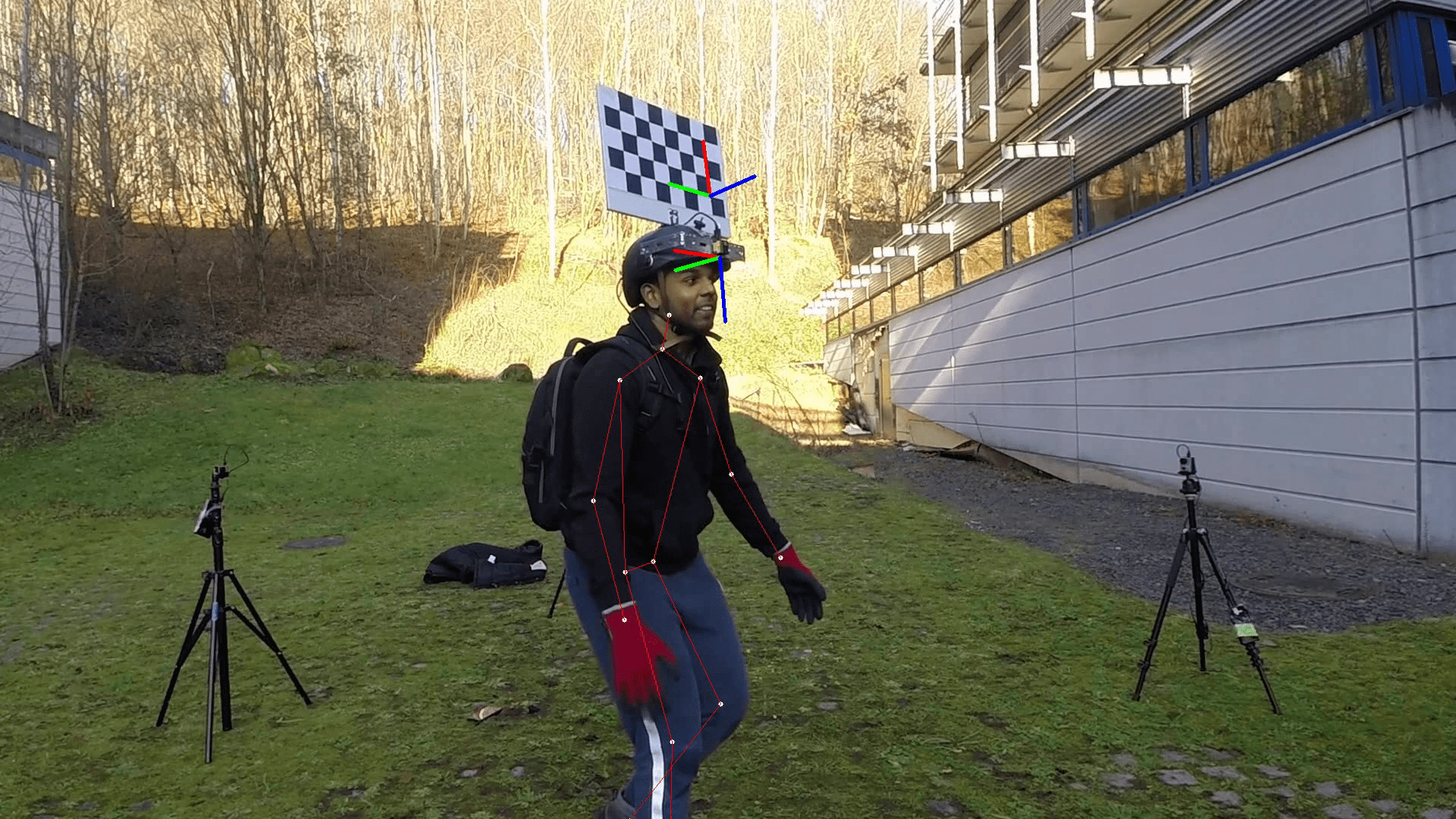}
\caption{
\textbf{Visualisation of the calibrated HMD and 3D human body pose.}
We employ a multi-view camera setup to simultaneously track the 3D human body pose and the position of a checkerboard in the world coordinate frame. 
The 3D poses obtained are subsequently projected onto the coordinate frame of the HMD. 
To establish the coordinate frame of the HMD, we determine a suitable transformation matrix that maps points from the checkerboard's coordinate frame to the HMD's coordinate frame. 
Given the known position of the checkerboard, this transformation matrix allows us to derive the egocentric 3D pose.
}
\label{fig:rgb_projection}
\end{figure}

We obtain the 3D human poses and SMPL~\citep{SMPL:2015} parameters within the world coordinate frame using the multi-view RGB camera setup (see Fig. \ref{fig:rgb_projection}). 
Subsequently, we apply the world-to-device transformation matrix $\mathbf{M}_{\text{WE}} \in \mathbb{R}^{4 \times 4}$  to convert these 3D human poses and SMPL parameters from the world coordinate frame to the HMD coordinate frame. 
Specifically, we use the following transformations:
\begin{equation}
    \label{eq:J3DWE}
    \mathbf{J} = \mathbf{M}_{\text{WE}} \cdot \mathbf{G}
\end{equation}
\begin{equation}
    \label{eq:SWE}
    \mathbf{S_E} = \mathbf{M}_{\text{WE}} \cdot \mathbf{S_W}
\end{equation}
Here, $\mathbf{G} \in \mathbb{R}^{16 \times 3}$ represents the world 3D human pose, $\mathbf{J} \in \mathbb{R}^{16 \times 3}$ represents the egocentric 3D human pose, $\mathbf{S_W} \in \mathbb{R}^{6890 \times 3}$ is the world SMPL mesh and $\mathbf{S_E} \in \mathbb{R}^{6890 \times 3}$ is the egocentric SMPL mesh.
Additionally, we derive the 2D egocentric joint coordinates, represented as $\mathbf{J_{2D}} \in \mathbb{R}^{16 \times 2}$, by projecting the egocentric 3D poses using the intrinsics of the event camera.

Also, we generate human body masks and visibility masks for each joint, in addition to obtaining the 3D human poses and SMPL parameters.
The joint visibility mask $V \in \{0, 1\}$ indicates whether a joint is visible or occluded from the egocentric view. 
We use \citet{blender_soft} to create the human body masks and the joint visibility masks.
We first set up a SMPL body model of the user and an egocentric virtual camera with the same intrinsic parameters and position as our real-world event camera.
To render the human body masks, we use Mist render layers in Blender's Cycles renderer.
Next, we obtain the joint visibility masks by shooting rays from the virtual camera to each 3D body joint.
When a ray intersects with the SMPL body for the first time, we query the nearest vertices of the intersection.
If the nearest vertices belong to the corresponding body part of the targeted 3D body joint, we mark that body joint as visible. 
Conversely, if the nearest vertices do not belong to the relevant body part, the 3D body joint is considered occluded.
Additionally, if a 3D joint is occluded, we also mark the corresponding 2D joint as occluded. 
The body parts are identified using the predefined human part segmentation mesh provided by \citet{SMPL:2015}.

\vspace{-0.2cm}

\section{Reconstructing Images from the Event Stream}
\label{appen:ee3dr_e2vid}
\begin{figure}[!htbp]
\centering
   \includegraphics[width=1\linewidth]{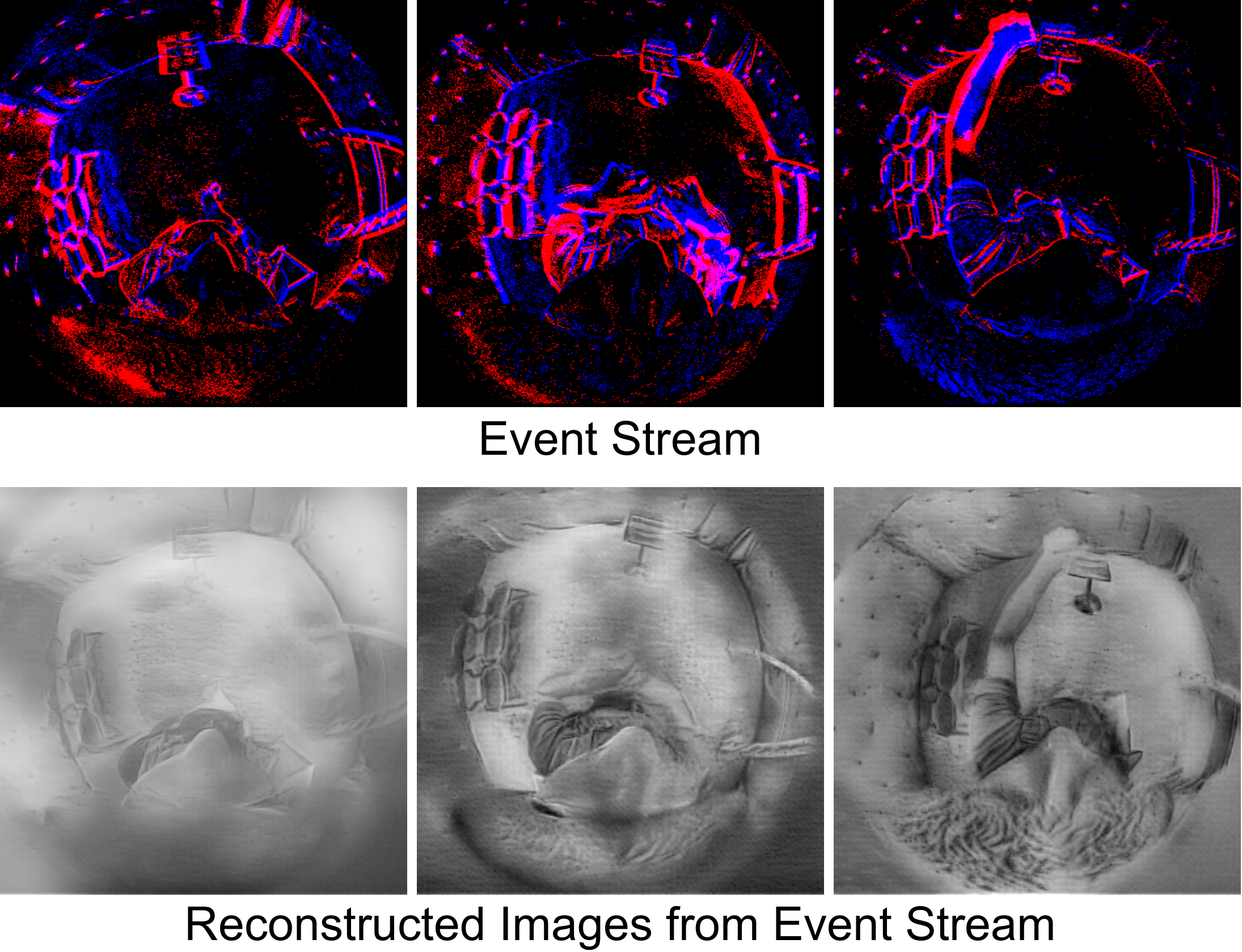}
   \caption{
   \textbf{Exemplary event streams and their corresponding image reconstructions.} 
    The reconstructed images lose significant details of the human body, especially when the motion of the human is minimal.
   }
\label{fig:E2VID_examples}
\end{figure}

We utilise E2VID~\cite{Rebecq19pami} to generate image reconstructions from the event stream. 
The frame duration (event window) is set to $20$~ms to align with the ground-truth frame timing of EE3D-R.
As shown in Fig.~\ref{fig:E2VID_examples}, the reconstructed images often exhibit artefacts, particularly in scenarios with minimal human motion. 
For instance, during low-motion actions such as walking (left part of Fig.~\ref{fig:E2VID_examples}), the reconstructed images fail to accurately capture the human figure. 
In contrast, during high-motion actions, such as punching (right part of Fig.~\ref{fig:E2VID_examples}), the reconstructed images can recover the human figure properly.
To ensure precise synchronization, each event window is aligned with the corresponding ground-truth frame number, maintaining consistency between the ground-truth 3D poses and the reconstructed images.

\vspace{-0.2cm}

\section{Synthetic Data Generation}
\label{appen:ee3ds_specs}

To simulate human motions as captured by event cameras, we linearly interpolated SMPL body parameters from the SURREAL dataset at a frequency of $480$Hz.
The dataset is created by generating RGB frames and human body masks through the Image and Mist render layers in Blender's Cycles renderer \citep{blender_soft}.
The 3D body joints used for training our EventEgo3D++ method, denoted as $\mathbf{J} = \{\mathbf{J}_1, \dots, \mathbf{J}_N\}$, where $\mathbf{J} \in \mathbb{R}^{16 \times 3}$, are derived from the SMPL body joints represented by $\mathbf{S} = \{\mathbf{S}_1, \dots, \mathbf{S}_N\}$, where $\mathbf{S} \in \mathbb{R}^{45 \times 3}$. Specifically, we map the joints as follows:
\begin{align*}
    \mathbf{G} &= \{ \mathbf{S}_{16}, \mathbf{S}_{13}, \mathbf{S}_{18}, \mathbf{S}_{20}, \mathbf{S}_{22}, \mathbf{S}_{17}, \mathbf{S}_{19}, \mathbf{S}_{21}, \\
    &\quad \mathbf{S}_{3}, \mathbf{S}_{6}, \mathbf{S}_{9}, \mathbf{S}_{12}, \mathbf{S}_{2}, \mathbf{S}_{5}, \mathbf{S}_{8}, \mathbf{S}_{11} \},
\end{align*}
where $\mathbf{S}_{i}$ denotes the $i$-th SMPL joint index from the set $\{1, 2, \dots, 45\}$. Each joint in $\mathbf{J}$ corresponds to a specific body part: the head, neck, right shoulder, right elbow, right wrist, left shoulder, left elbow, left wrist, right hip, right knee, right ankle, right foot, left hip, left knee, left ankle, and left foot, respectively.

\vspace{-0.2cm}

\section{Input representation}
\label{appen:input_repres}
We use the LNES representation~\citep{rudnev2021eventhands} to aggregate events over a time window without applying any temporal overlap. 
Nevertheless, we have conducted experiments using explicitly overlapping LNES frames with a $7$~ms temporal resolution matching our network's runtime performance of $140$~fps---which yields an MPJPE of $102.26$ and a PA-MPJPE of $75.62$ on the EE3D-R dataset. 
These results are nearly identical to our default setting using non-overlapping LNES frames (MPJPE: $102.15$, PA-MPJPE: $75.48$).

Furthermore, when using an overlapping configuration with a $1$~ms temporal resolution, we obtain an MPJPE of $100.54$ and a PA-MPJPE of $73.97$---corresponding to a $1.58\%$ reduction in MPJPE and a $2.00\%$ reduction in PA-MPJPE compared to our default configuration. 
However, since our network can only process frames at an effective rate of approximately $7$~ms per frame ($1000/140\approx7$~ms), this $1$~ms configuration is not feasible for real-time operation. 

\end{appendices}

\bibliography{bibliography}

\end{document}